\documentclass{article}

\usepackage{microtype}
\usepackage{graphicx}
\usepackage{subcaption}
\usepackage{booktabs} 

\usepackage{hyperref}
\usepackage{xurl}

\usepackage[preprint]{icml2026}

\usepackage{amsmath}
\usepackage{amssymb}
\usepackage{mathtools}
\usepackage{amsthm}

\definecolor{ciblue}{RGB}{100, 150, 240}
\definecolor{ciorange}{RGB}{250, 170, 80}
\newcommand{\ci}[1]{{\scriptsize $\pm#1$}}
\newcommand{\cib}[1]{{\color{ciblue}\scriptsize $\pm#1$}}
\newcommand{\cio}[1]{{\color{ciorange}\scriptsize $\pm#1$}}
\newcommand{\du}[1]{{\color{blue}\scriptsize $\uparrow#1$}}   
\newcommand{\dd}[1]{{\color{red}\scriptsize $\downarrow#1$}}

\renewcommand{\algorithmiccomment}[1]{\hfill$\triangleright$~#1}
\usepackage{multirow}

\usepackage[capitalize,noabbrev]{cleveref}

\theoremstyle{plain}
\newtheorem{theorem}{Theorem}[section]
\newtheorem{proposition}[theorem]{Proposition}
\newtheorem{lemma}[theorem]{Lemma}

\theoremstyle{definition}
\newtheorem{definition}[theorem]{Definition}

\theoremstyle{remark}
\newtheorem{remark}[theorem]{Remark}

\usepackage[textsize=tiny]{todonotes}

\icmltitlerunning{\texttt{Flat}-RLVR: Sharpness-Aware Reinforcement Learning with Verifiable Rewards for Reasoning Capacity in LLMs}

\begin{document}

\twocolumn[
  \icmltitle{\texttt{Flat}-RLVR: Sharpness-Aware Reinforcement Learning \\ with Verifiable Rewards for Reasoning Capacity in LLMs}

  \icmlsetsymbol{equal}{*}

   \begin{icmlauthorlist}
    \icmlauthor{Yeonbum Kim}{gsai}
    \icmlauthor{Hyun Kyu Lee}{gsai}
    \icmlauthor{Sung Whan Yoon}{gsai,ee}
  \end{icmlauthorlist}

  \icmlaffiliation{gsai}{Graduate School of Artificial Intelligence, Ulsan National Institute of Science and Technology, Ulsan, Republic of Korea}
  \icmlaffiliation{ee}{Department of Electrical Engineering, Ulsan National Institute of Science and Technology (UNIST), Ulsan, Republic of Korea}

  \icmlcorrespondingauthor{Sung Whan Yoon}{shyoon8@unist.ac.kr}

  \icmlkeywords{Distribution Sharpening, RLVR, Flatness, Reasoning Capability}

  \vskip 0.3in
]



\printAffiliationsAndNotice{}  

\begin{abstract}
Reinforcement learning with verifiable rewards (RLVR), particularly through group-relative policy optimization (GRPO), has emerged as a central paradigm for the post-training of large language models (LLMs). While RLVR is highly effective at enhancing the reasoning capabilities of LLMs, it commonly induces \textit{distribution sharpening}, whereby the policy distribution becomes excessively peaked and rare-yet-correct trajectories are progressively suppressed, thus diminishing sample diversity. To this end, we promote flat-reward optima in RLVR to smooth the policy distribution.
We theoretically link flat reward RLVR to reduced distribution sharpening, and empirically show improved sample diversity while maintaining single-attempt accuracy. Our method, called \texttt{Flat}-RLVR, searches for flat reward regions via sharpness-aware minimization during LLM post-training, combining this with GRPO variants to achieve better trade-offs between sampling efficiency and reasoning capacity.
\end{abstract}

\section{Introduction}
Reinforcement learning with verifiable rewards (RLVR), especially through \emph{group-relative} methods, has emerged as a predominant methodology for large language model (LLM) post-training~\cite{deepseekmath,Tulu,deepseek-r1}. Its prominence stems from two complementary strengths: first, rewarding only the correctness of the final answer is sufficient to elicit valid reasoning behaviors~\cite{deepseek-r1,spurious}; and second, the group-relative method eliminates the value network required by proximal policy optimization (PPO)~\cite{ppo}, substantially reducing the memory burden of maintaining a critic of comparable size to the policy~\cite{deepseekmath}. This critic-free design---first realized as group-relative policy optimization (GRPO)~\cite{deepseekmath} and subsequently extended in its variants---has rapidly become the \emph{de facto} standard for RLVR.

\begin{figure}[t]
  \vskip 0.2in
  \begin{center}
  \centerline{\includegraphics[width=\columnwidth]{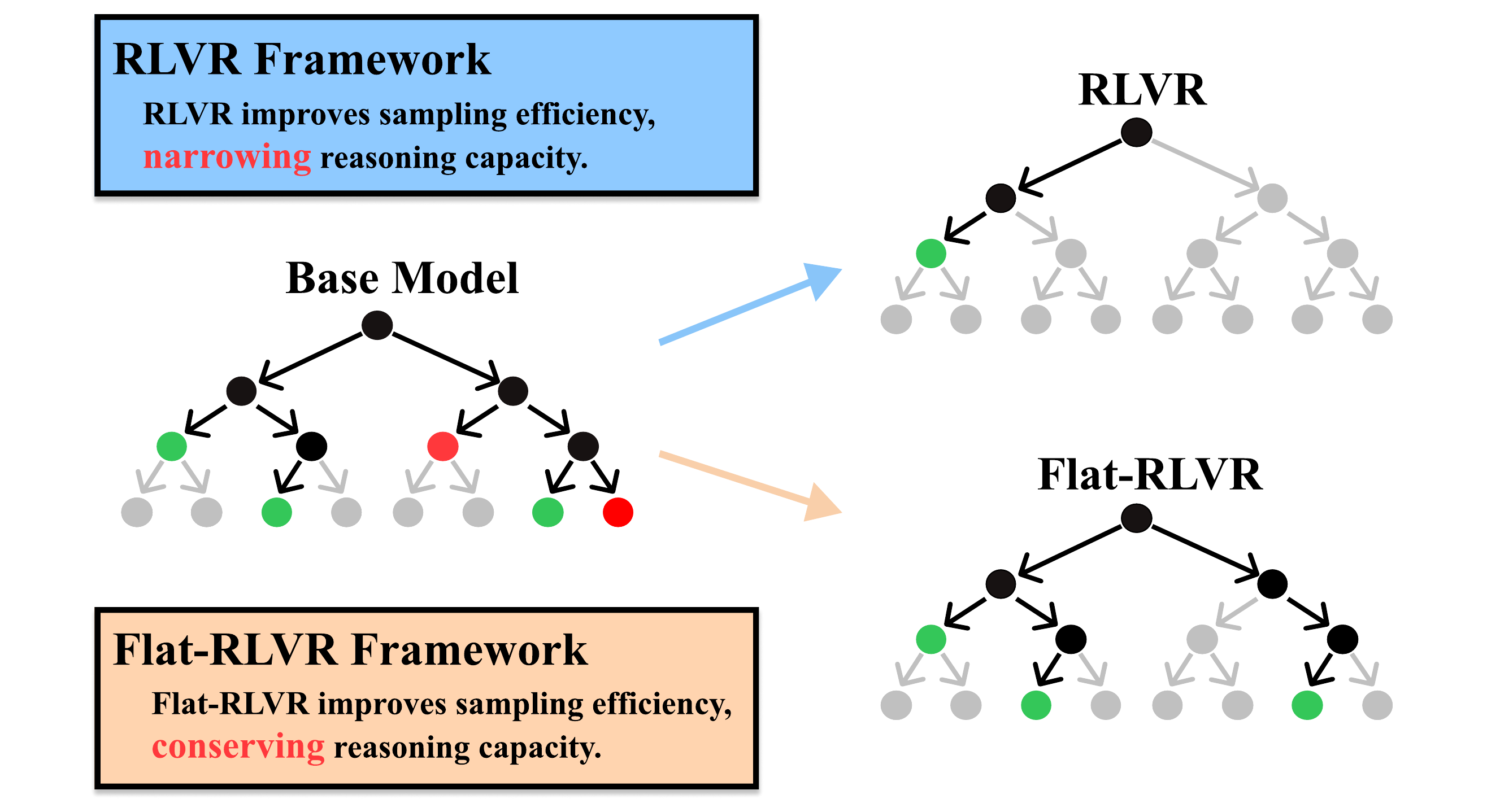}}
  \caption{RLVR and \texttt{Flat}-RLVR on LLM reasoning capacity. \textcolor{gray}{Gray} nodes indicate paths that are unlikely to be sampled by the model, \textcolor{black}{black} nodes indicate paths that are likely to be sampled, i.e., each path consists of a sequence of outputs, later formulated by $o_j$, where $j$ is the path index. \textcolor{green}{Green} nodes indicate correct solutions, and \textcolor{red}{red} nodes indicate incorrect. 
  }
   \label{fig:main_concept}
\end{center}
\end{figure}

However, a critical doubt has been raised, stemming from empirical observations that GRPO impairs its efficacy when evaluated across multiple samples by checking whether any one succeeds; so-called Pass@$K$ performance metric~\cite{deepseekmath}. Specifically, base LLM models without post-training even surpass their RLVR-tuned counterparts under Pass@$K$ with sufficiently large $K$~\cite{doesRL}. 
This misgiving has recently raised a hypothesis that \textit{RLVR merely improves sampling efficiency at the expense of the model's underlying reasoning capacity driven by LLM's sample diversity}~\cite{reasoning_boundary, echo_chamber, surprising}. Further evidence reinforces this view: RLVR has been shown to amplify only responses that already carry high probability mass while neglecting the long tail of rare-yet-correct samples, thereby reducing sample diversity~\cite{unlikely}; and to concentrate exclusively on regions already solvable by the base model, failing to acquire novel capabilities~\cite{doesRL}. Taken together, these findings further support the interpretation that the current RLVR mechanism induces \textit{distribution sharpening}---a phenomenon in which the model reinforces only the easier, already-solved instances and loses its exploratory capacity.

This is particularly problematic in light of the emerging paradigm of inference-time compute scaling, which leverages sample diversity for effective LLM deployment~\cite{self_consis,monkey, inference-aware, guide_sampling, sample_diver}. These strategies operate on a simple principle---generate many candidate solutions and rely on a verifier to identify the correct ones---and are particularly well aligned with RLVR's domains of mathematics and code, where deterministic verifiers naturally enable reliable selection~\cite{gsm8k, pass, monkey}.
In this regard, distribution sharpening from RLVR, which is observed to trade output diversity for efficiency, directly undermines LLM's benefits of inference-time computation. Figure \ref{fig:main_concept} presents LLM-generated output trajectories (i.e., reasoning paths), the base model covers sometimes-incorrect but wide variety of answers, but RLVR \textit{narrows} the path to a few answers discarding diversity.

In response to this growing need for sample diversity, we apply reinforcement learning (RL) to flatten the reward surface of the RLVR framework, thereby smoothing policy distributions and diversifying the generated candidate samples.
Recently, it has been shown that a flatter maximum on the reward surface implies a robust policy against uncertain perturbations on actions, states, and rewards~\cite{ppo+sam}.
Motivated by the flat RL concept, our key insight is that, rather than updating the LLM toward steepest loss descent, steering it to flatten the surface may smooth the distribution sharpening.

In this paper, we find that sharpness-aware minimization (SAM) variants~\cite{sam,LookSAM,gcsam} applied to RL optimization within the RLVR paradigm indeed mitigate distribution sharpening by yielding a flatter reward 
(as illustrated in Figure~\ref{fig:main_concept}).
In addition, we formalize how the flat-search optimization mathematically lifts suboptimal candidates by smoothing the reward surface. Our contributions are highlighted as follows:
\textbf{i)} We propose a novel strategy to alleviate the distribution sharpening of GRPO by borrowing the concept of flat-minima search. \textbf{ii)} We provide a theoretical analysis of how a flat-inspired GRPO elevates suboptimal candidates, leading to the smoothed output distribution. \textbf{iii)} We suggest a \textbf{\texttt{Flat}-RLVR} framework by integrating SAM into GRPO methods. \texttt{Flat}-RLVR, instantiated by combining with GRPO~\cite{deepseekmath}, DAPO~\cite{dapo}, and CISPO~\cite{cispo}, consistently improves Pass@$256$ mathematical reasoning capacity while preserving accuracy with our LLM post-training adaptations.

\section{Related Work}
\subsection{Group-Relative RLVR}
Reinforcement learning with verifiable rewards (RLVR) trains a large language model on binary rewards provided by deterministic verifiers, eliciting strong reasoning behavior without human-preference data~\cite{deepseek-r1,Tulu,spurious}. Within this paradigm, group-relative methods have become dominant, yielding computation benefits by normalizing rewards across a group of rollouts sampled from the same prompt and thereby removing the need for a separate value network~\cite{deepseekmath}. The original instance, GRPO~\cite{deepseekmath}, has since been extended in several directions. DAPO~\cite{dapo} introduces asymmetric clipping bounds that allow low-probability tokens to bring larger upward updates, together with dynamic sampling that discards groups whose advantages collapse to zero. CISPO~\cite{cispo} clips the importance-sampling weight to prevent dropping tokens that carry essential information.

Our work integrates a sharpness-aware optimization perspective into the reinforcement learning loop, showing a general applicability across the group-relative RLVR family.

\subsection{Distribution Sharpening in RLVR}
While the distribution sharpening phenomenon of RLVR is now well documented~\cite{doesRL, reasoning_boundary, echo_chamber}, existing mitigation strategies all intervene at the level of the reward or token distribution. F-GRPO~\cite{fgrpo} applies a Focal Weight to down-weight easy prompts, the unlikeliness reward~\cite{unlikely} amplifies signals from rare correct trajectories, and clipping-based variants such as DAPO~\cite{dapo} relax token-level updates to preserve low-probability tokens. 

We address distribution sharpening via \emph{optimization geometry itself}, biasing updates toward flat regions of the loss landscape via SAM---a direction is novel and orthogonal to the above approaches.

\subsection{Advances in SAM}
Sharpness-aware minimization (SAM) has been improved as a primary driver of flat-minima search by addressing its drawbacks. The first concern is its sensitivity to parameter re-scaling: ASAM~\cite{asam} addresses this by introducing a scaling operator that adaptively adjusts the perturbation radius according to parameter magnitudes. The second issue is SAM's vulnerability to noisy gradients, particularly during the early stages of training; GCSAM~\cite{gcsam} mitigates this through Gradient Centralization, which removes the mean of the gradient to yield a more reliable perturbation direction. The third concern is the computational overhead of SAM; LookSAM~\cite{LookSAM} amortizes this cost by computing the SAM update only periodically, while Momentum-SAM~\cite{momentum_sam} integrates Nesterov Accelerated Gradient (NAG) into SAM to directly improve computational efficiency. Lastly, the observation that SAM does not invariably converge to flat minima leads to GSAM~\cite{GSAM} adopting a surrogate gap formulation for the sharpness metric.

Our work adopts the idea of several SAM-based variants to build the \texttt{Flat}-RLVR framework; applying the SAM update periodically, inspired by LookSAM~\cite{LookSAM} and Gradient Centralization, inspired by GCSAM~\cite{gcsam}. Notably, \texttt{Flat}-RLVR is not limited to a certain type of SAM approaches.

\subsection{Flat-Minima and Generalization}
Flat minima have been studied as being linked to improved generalization~\citep{flat_minima, large_batch_training} emphasized in domain generalization~\cite{swad} and adversarial robustness~\cite{flat_minima_adversarial} with underlying principles~\cite{sam_duality,weebum}.
Researchers have extended flat-minima toward various learning frameworks, including generative modeling~\citep{flat_minima_diffusion}, federated learning~\citep{fedsam,FedGF}, and reinforcement learning (RL)~\citep{ppo+sam}. 
Notably, RL with flatter reward yields more robust policies to action, state, and reward uncertainties \citep{ppo+sam}. Such flat reward insight makes us anticipate potential benefits in RLVR, but it is nontrivial to identify how flatness-aware RL steers LLM behavior. 

Our work is the \emph{first} attempt to employ SAM in LLM post-training, revealing its benefits in relieving distribution sharpening in RLVR.

\section{Preliminaries}
\subsection{Group Relative Policy Optimization}
GRPO~\cite{deepseekmath} removes the value function, then instead calculates advantages relative to a group of sampled rollouts. For a prompt $q\sim Q$ with $G$ rollouts $\{o_i\}^G_{i=1}$, and rewards $\{R_i\}^G_{i=1}$, the group-relative advantage for $i$-th rollout is
\begin{equation}
\label{eq:grpo_advantage}
    \hat{\mathcal{A}}_i = \frac{R_i - \bar{R}}{\sigma_R + \varepsilon},
\end{equation}
where $\bar{R} := \frac{1}{G}\sum_{i}R_i$, $\sigma_R=\textrm{std}(\{R_j\})$ and $\varepsilon>0$.

Let $o_i = (y_{i,1}, ..., y_{i,|o_i|})\in O$ denote the token sequence for rollout $i$ computed by the old policy $\pi_{\theta_{\text{old}}}(O|q)$ from a prompt $q$. GRPO optimizes policy $\pi_{\theta}$ to maximize the following clipped surrogate objective based on group-relative advantage $\hat{\mathcal{A}}_{i,\tau}$:
\begin{multline}
\label{eq:grpo_objective}
    J_{\text{GRPO}}(\theta) = \mathbb{E}_{q \sim P(Q), \{o_i\}_{i=1}^G \sim \pi_{\theta_{\text{old}}}(O|q)} \bigg[   \\
    \frac{1}{G} \sum_{i=1}^{G}  \frac{1}{|o_i|} 
\sum_{\tau=1}^{|o_i|}\min\Big\{\frac{\pi_\theta(o_{i,\tau} \mid q, o_{i,<\tau})}
{\pi_{\theta_{\text{old}}}(o_{i,\tau} \mid q, o_{i,<\tau})}
\hat{\mathcal{A}}_{i,\tau},    \\
\mathrm{clip}\big(\frac{\pi_\theta(o_{i,\tau} \mid q, o_{i,<\tau})}
{\pi_{\theta_{\text{old}}}(o_{i,\tau} \mid q, o_{i,<\tau})},
1 - \epsilon, 1 + \epsilon
\big)
\hat{\mathcal{A}}_{i,\tau}\Big\}\bigg],    \\
\end{multline}
where $o_{i,\tau}=y_{i,\tau}$, $o_{i,<\tau}=\{y_{i,1},\cdots,y_{i,\tau-1}\}$, and $\hat{\mathcal{A}}_{i,\tau}$ is the $i$-th rollout advantage given with a set of partial sequences $\{o_{i,\leq\tau}\}_{i=1}^{G}$. In RLVR, all tokens within the same rollout share an identical advantage value; $\hat{\mathcal{A}}_{i,\tau} = \hat{\mathcal{A}_{i}}$ for all $\tau$. $\mathrm{clip}(\cdot,a,b)$ clips input to be in interval $(a,b)$, where $a<b\in\mathbb{R}$.

We drop the Kullback-Leibler regularization term adopted by GRPO by following recent findings that the regularization hinders explorations in reasoning~\cite{klterm}.

\begin{figure*}[t]
  \includegraphics[width=\linewidth]{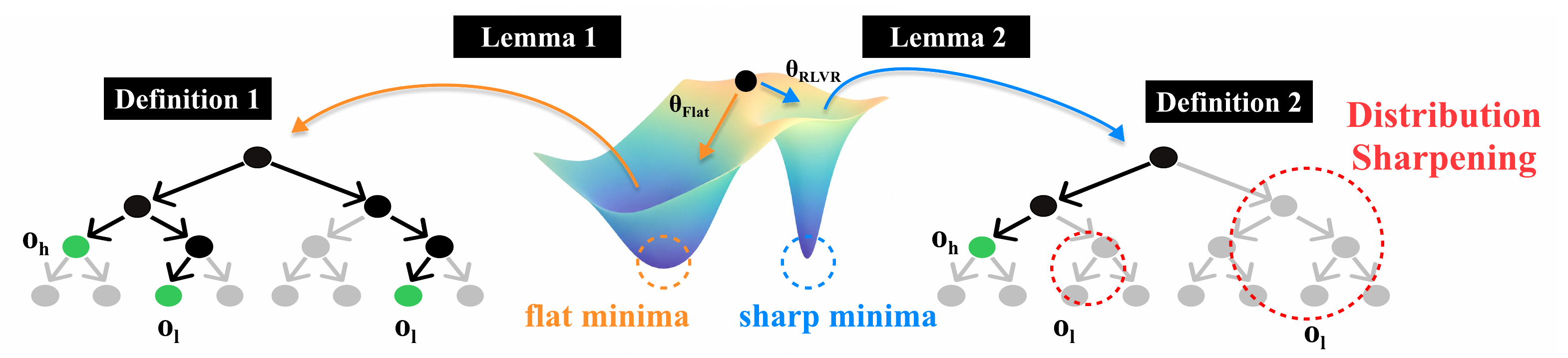}
  \centering
  \caption{Illustration of the loss landscape and reasoning capacity of RLVR and \texttt{Flat}-RLVR. 
  \texttt{Flat}-RLVR converges to a flatter minimum in the loss landscape by jointly optimizing for loss reduction and sharpness minimization, whereas RLVR is drawn to a sharper minimum driven solely by reward maximization. 
  As shown in the trees (see Figure~\ref{fig:main_concept}), \texttt{Flat}-RLVR explores a broader reasoning tree, while RLVR collapses to a narrow set of paths.}
   \label{fig:formula}
\end{figure*}

\subsection{Sharpness-Aware Minimization}
SAM~\cite{sam} seeks a minimum with both small loss and curvature, biasing optimization toward flat regions of the loss landscape. Formally, SAM solves the following min-max objective:
\begin{equation}
\label{eq:sam_objective}
    \underset{\theta}{\min} \ \underset{\|\epsilon\|_2 \leq \rho}{\max} \ \mathcal{L}(\theta + \epsilon),
\end{equation}
where $\mathcal{L}(\theta)$ is the loss function parametrized by $\theta$, and $\epsilon$ is a parameter perturbation constrained to lie within an $\ell_2$ ball of radius $\rho>0$. The inner maximization admits a first-order approximation so that the worst-case perturbation becomes
\begin{equation}
\label{eq:sam_perturbation}
    \epsilon^* = \rho{\nabla_{\theta}\mathcal{L}(\theta)}/{\lVert\nabla_{\theta}\mathcal{L}(\theta)\rVert}.
\end{equation}
By minimizing the worst-case loss within this perturbation neighborhood, SAM penalizes sharp minima and encourages convergence to flat ones.

\section{Analysis on Flatness in RLVR}
\label{sec:analysis}
Based on the GRPO and SAM objectives, we analyze how a sharpness-aware RLVR, so-called \texttt{Flat}-RLVR, affects distribution sharpening over LLM outputs. Without loss of generality, we use a \textit{local} sequence-level unclipped surrogate induced by a fixed rollout group~\cite{sequence_level}. For a prompt $q$, let $J_q(\theta)$ denote the local RLVR objective, which drops the expectation and clipping in Eq.~\eqref{eq:grpo_objective}.
Set $\mathcal L_q(\theta)=-J_q(\theta)$; maximizing RLVR objective is equivalent to minimizing $\mathcal L_q$.
Figure~\ref{fig:formula} previews the key components of our analysis that follows. 

\begin{definition}[$\rho$-\texttt{Flat}-RLVR]
\label{def1}
Given prompt $q$ and the computed local sequence-level RLVR loss $\mathcal L_q(\theta)$, let $\rho$-sharpness of $\mathcal L_q$ at parameter $\theta$ be
\begin{equation}
\operatorname{Sharp}_{\rho}(\theta;q)=\max_{\|\epsilon\|_2\le\rho} \left[\mathcal L_q(\theta+\epsilon)-\mathcal L_q(\theta)\right],
\label{eq:rho_sharpness}
\end{equation}
where $\epsilon$ is a parameter perturbation and $\|\cdot\|_2$ is Euclidean norm.
Then, the $\rho$-\texttt{Flat} RLVR objective is defined as the expected sharpness-aware loss:
\begin{multline}
 \min_{\theta}\mathbb E_{q\sim Q} \left[\mathcal L_q(\theta) + \operatorname{Sharp}_{\rho}(\theta;q)\right] \\ = \min_{\theta} \mathbb E_{q\sim Q} \left[\max_{\|\epsilon\|_2\le\rho} \mathcal L_q(\theta+\epsilon)\right].    
\label{eq:flat_obj}
\end{multline}
\end{definition}

Following SAM, \texttt{Flat}-RLVR optimizes Eq.~\eqref{eq:flat_obj} by a two-step update at parameters $\theta_t$, where $t$ is an update index: it first ascends to the worst-case perturbation $\epsilon_t^* = \rho\,\nabla_\theta\mathcal L_q(\theta_t)/\|\nabla_\theta\mathcal L_q(\theta_t)\|_2$, then descends using the gradient at the perturbed point,
\begin{equation}
\label{eq:parameter_update}
    \theta_{t+1} = \theta_t - \alpha\,\nabla_\theta\mathcal L_q(\theta_t+\epsilon_t^*).
\end{equation}
The perturbation $\epsilon_t^*$ is what distinguishes \texttt{Flat}-RLVR from the standard update, which cannot account for loss sensitivity to perturbations.

\begin{definition}[Correct-set Distribution Sharpening]
\label{def2}
Given prompt $q$ and policy distribution $\pi_\theta(\cdot|q)$ over outputs $o\in O$, let $\mathcal C_q=\{o:R(o)=1\}$ be a verifier-correct set, whose elements show $1$-reward.
For two correct outputs $o_h, o_l\in\mathcal C_q$ satisfying
\begin{equation}
    \pi_{\theta_{\mathrm{old}}}(o_h|q) > \pi_{\theta_{\mathrm{old}}}(o_l|q),
\end{equation}
let us formulate their correct-set log-probability gap at parameter $\theta$ as
\begin{equation}
    D_{h,l}(\theta;q) = \log\frac{\pi_\theta(o_h|q)}{\pi_\theta(o_l|q)}.
\end{equation}
An update from $\theta_{\mathrm{old}}$ to $\theta_{\mathrm{new}}$ is defined to sharpen the correct-set distribution on $(o_h, o_l)$ if
\begin{equation}
    D_{h,l}(\theta_{\mathrm{new}};q) - D_{h,l}(\theta_{\mathrm{old}};q) > 0.
\end{equation}
\end{definition}

Here, $o_h$ denotes a correct reasoning path with higher probability under the old policy, while $o_l$ denotes a lower-probability but still correct reasoning path.
We now analyze how the policy update via $\rho$-\texttt{Flat}-RLVR affects the confidence gap $D_{h,l}$.

\begin{lemma}[$\rho$-\texttt{Flat}-RLVR One-Step Decomposition]
\label{lem:Flat-RLVR}
    The one-step update of \texttt{Flat}-RLVR from Definition~\ref{def1} admits the following local expansion:
\begin{equation}
    \label{eq:lemma_decomp}
        \theta_{t+1}^{\texttt{Flat}} = \theta_t + \alpha\Big(\underbrace{g_t}_{\textnormal{Score Part}} \underbrace{-\,\rho\,\frac{M_t g_t}{\|g_t\|_2}}_{\textnormal{Flatness Part}}\Big) + R_t.
\end{equation}
Here, for the sequence-level score is $s_{i,t} = \nabla_\theta\log\pi_\theta(o_i|q)|_{\theta_t}$ over rollouts $\{o_i\}_{i=1}^G$, $g_t = \frac{1}{G}\sum_{i=1}^G \hat{\mathcal A}_i s_{i,t}$, $M_t = \frac{1}{G}\sum_{i=1}^G \hat{\mathcal A}_i s_{i,t} s_{i,t}^\top$.

The residual $R_t = -\alpha\rho\,B_t g_t/\|g_t\|_2 - \alpha R_T$ collects the log-policy Hessian term $B_t = \frac{1}{G}\sum_i \hat{\mathcal A}_i\nabla_\theta^2\log\pi_\theta(o_i|q)$ and the higher-order Taylor residual $R_T$.
\end{lemma}

The `Score Part' is the standard RLVR update, which increases the likelihood of positively advantaged rollouts along the reward-improving direction $g_t$. The `Flatness Part', scaled by $\rho$, opposes that direction through the curvature operator $M_t$, dampening over-amplification along $g_t$. Under a sequence-level neural tangent kernel (NTK) approximation~\cite{NTK_linear}, $B_t$ cancels by the zero-mean property of group-relative advantages, so the residual is negligible at leading order. The detailed derivation is provided in Appendix~\ref{app:proof_lemma1}.

\begin{lemma}[Distribution Sharpening One-step Decomposition]
\label{lem:correct_gap}
    For a correct pair $(o_h, o_l)$ in Definition~\ref{def2}, let $\Delta s_t := s_{h,t}-s_{l,t}$ with $s_{h,t} = \nabla_\theta\log\pi_\theta(o_h|q)|_{\theta_t}$ and $s_{l,t} = \nabla_\theta\log\pi_\theta(o_l|q)|_{\theta_t}$.
    For a one-step parameter update $\Delta\theta_t = \theta_{t+1} - \theta_t$, define the first-order correct-set gap change as
    \begin{multline}
    \label{eq:first_order_correct_set}
        \mathfrak D_t(\Delta\theta_t) := D_{h,l}(\theta_t+\Delta\theta_t;q) - D_{h,l}(\theta_t;q) \\ \approx \Delta s_t^\top\Delta\theta_t.
    \end{multline}
    Then the $\rho$-\texttt{Flat}-RLVR update satisfies
    \begin{equation}
        \label{eq:lemma_final_eq}
        \mathfrak D_t(\Delta\theta_t^{\texttt{Flat}}) = \mathfrak D_t(\Delta\theta_t^{\mathrm{GRPO}}) - \alpha\rho C_t + r_t,
    \end{equation}
    where $\Delta\theta_t^{\mathrm{GRPO}}=\alpha g_t$, $C_t = (\Delta s_t^\top M_tg_t) / \|g_t\|_2$, and $r_t=\Delta s_t^\top R_t.$
\end{lemma}

Here, $C_t>0$ indicates that the flatness correction reduces the GRPO-induced growth of the correct-set log-probability gap between $o_h$ and $o_l$. The proof is provided in Appendix~\ref{app:proof_lemma2}.

\begin{proposition}[\texttt{Flat}-RLVR Mitigates Correct-set Distribution Sharpening]
\label{prop:mitigation}
Along a training trajectory, suppose that the cumulative flatness correction dominates the residual:
\begin{equation}
    \sum_{t=0}^{T-1}\alpha\rho C_t \ge \sum_{t=0}^{T-1} r_t.
\end{equation}
Then the cumulative correct-set gap growth induced by $\rho$-\texttt{Flat}-RLVR is no larger than that induced by the corresponding GRPO score updates:
\begin{equation}
    \sum_{t=0}^{T-1}\mathfrak D_t(\Delta\theta_t^{\texttt{Flat}}) \le \sum_{t=0}^{T-1}\mathfrak D_t(\Delta\theta_t^{\mathrm{GRPO}}).
\end{equation}
\end{proposition}
Proposition~\ref{prop:mitigation} implies that $\rho$-\texttt{Flat} RLVR mitigates correct-set distribution sharpening relative to standard RLVR under the assumptions of our analysis. Proof is in Appendix~\ref{app:proof_proposition}.

\begin{remark}[Incorrect-output penalization]
Proposition~\ref{prop:mitigation} does not imply that \texttt{Flat}-RLVR raises the probability of incorrect outputs. For an incorrect output with a negative advantage, the same flatness correction term in Lemma~\ref{lem:Flat-RLVR} remains aligned with probability suppression. Thus, \texttt{Flat}-RLVR can reduce over-concentration among correct reasoning paths while still preserving reward separation between correct and incorrect outputs.
\end{remark}

\begin{algorithm}[tb]
  \caption{\texttt{Flat}-RLVR}
  \label{alg:sagrpo}
  \begin{algorithmic}[1]
    \REQUIRE Initial parameters $\theta_{0}$; perturbation radius $\rho$; learning rate $\alpha$; SAM period $n$; group size $G$; total steps $T$
    \FOR{$t = 1, \ldots, T$}
      \STATE \textit{// Rollout and advantage estimation}
      \STATE Sample $G$ completions $\{a_{1}, \ldots, a_{G}\} \sim \pi_{\theta_{\mathrm{old}}}(\cdot \mid q)$ \label{alg:line3}
      \STATE Compute group-relative advantages
        \[
          \hat{A}_{i} = \frac{r_{i} - \mathrm{mean}(\{r_{j}\})}{\mathrm{std}(\{r_{j}\})}
        \]
      \STATE Compute GRPO loss $\mathcal{L}(\theta)$
      \STATE $g_{\theta} \gets \nabla_{\theta} \mathcal{L}(\theta)$
      \IF[SAM step]{$t \bmod n = 0$}
        \STATE $g_{\theta} \gets g_{\theta} - \mu(g_{\theta})$ \COMMENT{GC (ascent)}
        \STATE $\rho_{t} \gets \mathrm{anneal}(\rho, t) \cdot \bigl(\hat{p}_{\mathrm{mean}}\bigr)^{\gamma}$ \COMMENT{Cosine + Focal}
        \STATE $\epsilon^{*} \gets \rho_{t}\,g_{\theta} / \lVert g_{\theta} \rVert_{2}$
        \STATE $\theta \gets \theta + \epsilon^{*}$ \COMMENT{ascend}
        \STATE $g_{\theta+\epsilon^{*}} \gets \nabla_{\theta} \mathcal{L}(\theta + \epsilon^{*})$ \label{alg:line12}
        \STATE $g_{\theta+\epsilon^{*}} \gets g_{\theta+\epsilon^{*}} - \mu(g_{\theta+\epsilon^{*}})$ \COMMENT{GC (descent)}
        \STATE $\theta \gets \theta - \epsilon^{*}$ \COMMENT{restore}
        \STATE $\theta \gets \theta - \alpha\,g_{\theta+\epsilon^{*}}$
      \ELSE[Standard GRPO step]
        \STATE $\theta \gets \theta - \alpha\,g_{\theta}$
      \ENDIF
    \ENDFOR
  \end{algorithmic}
  \vspace{2pt}
  {\color{gray}\small $\triangleright$ \textit{Line 9: Annealing is used with Focal Weight for $\rho$ scheduling. $\hat{p}_{\textrm{mean}}$ is an inverse prompt-difficulty factor (Appendix~\ref{app:focal_weight}).}}
\end{algorithm}

\begin{figure*}[t]
  \centering
  \begin{subfigure}[t]{0.245\linewidth}
    \includegraphics[width=\linewidth]{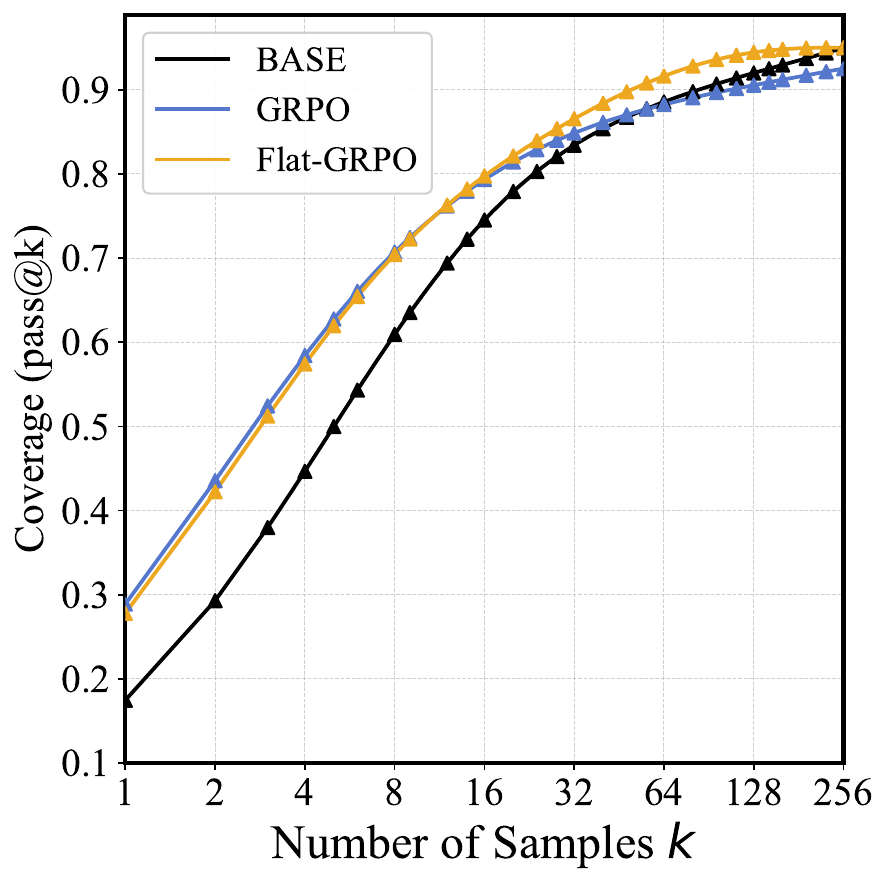}
    \caption{AMC23}
    \label{fig:amc23}
  \end{subfigure}
  \hfill
  \begin{subfigure}[t]{0.245\linewidth}
    \includegraphics[width=\linewidth]{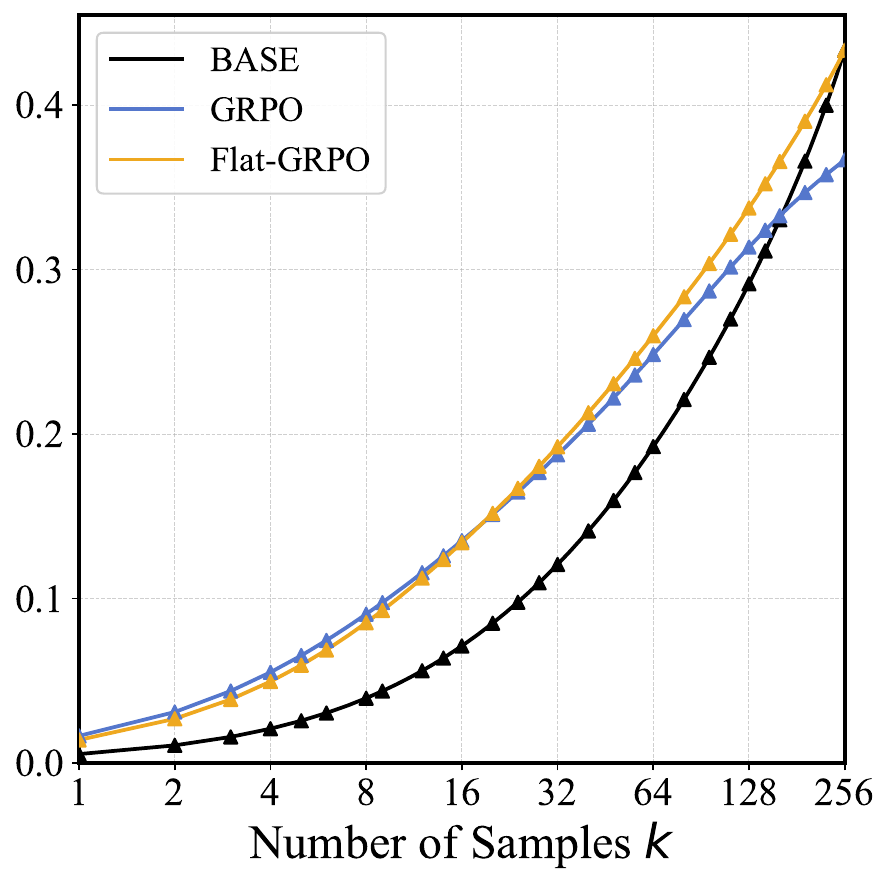}
    \caption{AIME24}
    \label{fig:aime24}
  \end{subfigure}
  \hfill
  \begin{subfigure}[t]{0.245\linewidth}
    \includegraphics[width=\linewidth]{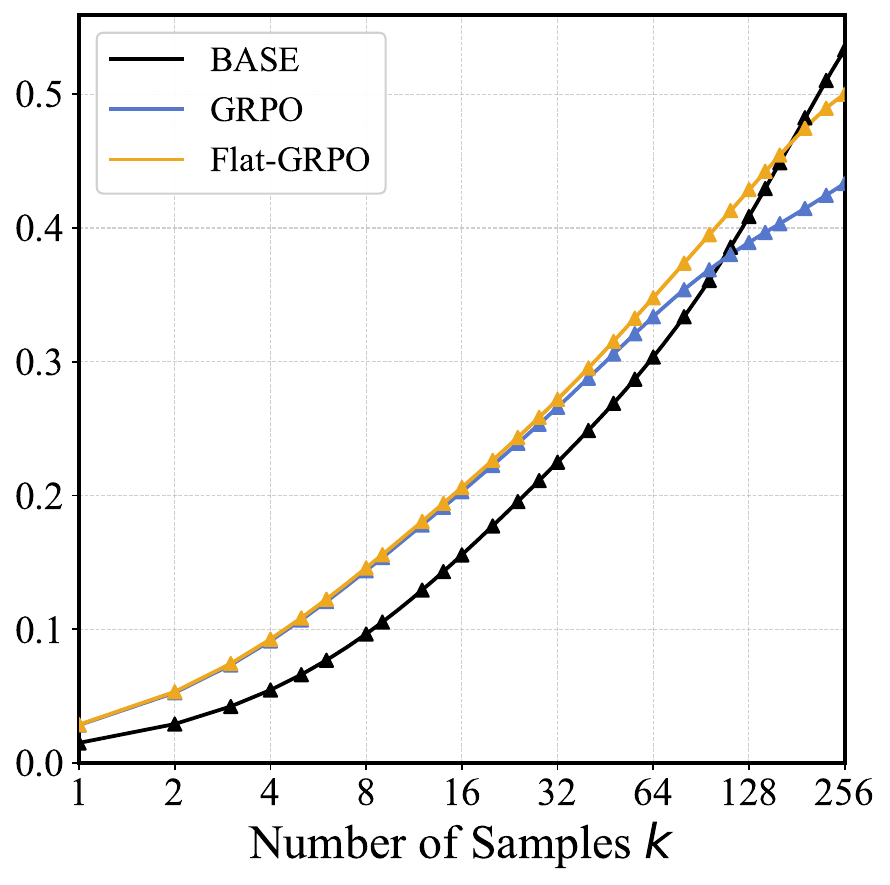}
    \caption{AIME25}
    \label{fig:aime25}
  \end{subfigure}
  \hfill
  \begin{subfigure}[t]{0.245\linewidth}
    \includegraphics[width=\linewidth]{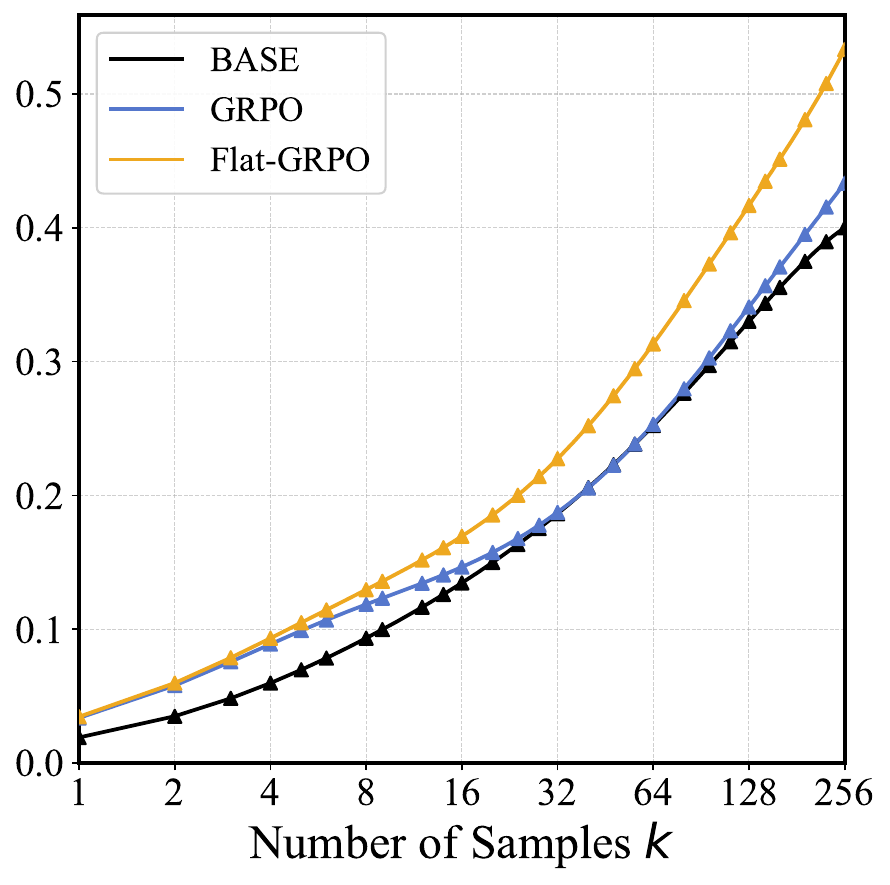}
    \caption{AIME26}
    \label{fig:aime26}
  \end{subfigure}
  \caption{Pass@K curves of BASE model ((a), (c), and (d)~Qwen2.5-Math-1.5B and (b)~Llama-3.2-1B-Instruct), GRPO, and \texttt{Flat}-GRPO (ours) on in-domain mathematical reasoning benchmarks, traced from $K=1$ to $K=256$. 
  All Pass@$K$ curves per-BASE/benchmark are provided in Figure~\ref{fig:all_curves}.}
  \label{fig:main}
\end{figure*}

\section{Proposed Method: \texttt{Flat}-RLVR}
\label{methods}

\subsection{\texttt{Flat}-RLVR}
Grounded in our theoretical findings that flatness-seeking optimization alleviates the distribution sharpening in RLVR, we construct the \texttt{Flat}-RLVR framework, which integrates SAM into group-relative RLVR. The \texttt{Flat}-RLVR objective is:
\begin{equation}
    \min_{\theta} \mathbb E_{q\sim Q} \left[\max_{\|\epsilon\|_2\le\rho} \mathcal L_q(\theta+\epsilon)\right].
\end{equation}

Specifically, for each prompt $q$, the current policy $\pi_{\theta_{\mathrm{old}}}$ generates $G$ completions, and the corresponding rewards are normalized within the group to yield the advantage estimator $\hat{A}_{i}$. This is the standard GRPO procedure and serves as the baseline for our SAM-based modifications.

First, we calculate the objective function $\mathcal{L}_q(\theta_t)$ and gradient $\nabla_\theta\mathcal{L}_{q}(\theta_t)$. Then, the ascent step for finding the worst perturbation is calculated by
\begin{equation}
    \epsilon_t^* = \rho\,\nabla_\theta\mathcal L_q(\theta_t)/\|\nabla_\theta\mathcal L_q(\theta_t)\|_2,
\end{equation}
The parameters are perturbed toward $\theta + \epsilon_t^*$, and the descent step for flat-minima is computed by
\begin{equation}
    \theta_{t+1} = \theta_t - \alpha\,\nabla_\theta\mathcal L_q(\theta_t+\epsilon_t^*)
\end{equation}
for updating policy, which corresponds to Eq~\eqref{eq:parameter_update}.

\subsection{Adaptations for LLM post-training}
\texttt{Flat}-RLVR operates under the SAM paradigm, and incorporates the following adaptations and advances tailored for LLM post-training.

\paragraph{Scalability.}
SAM induces one more gradient computation to obtain the perturbation.
We adopt SAM not for all steps, but for every $n$ steps by borrowing the existing technique~\cite{LookSAM}.
While this prior work only treats $n$ as a scalability knob, our adoption in LLM post-training introduces a new dimension: the period $n$ governs the effective degree of SAM regularization, and consequently impacts the Pass@$K$ trade-off. We provide further analysis of scalability on \texttt{Flat}-RLVR in Section~\ref{exp:periodic_sam}.

\paragraph{Perturbation radius.}
We adopt the prompt-difficulty-aware strategy of F-GRPO with Focal Weight~\cite{fgrpo}, which decays the advantage with difficulty. 
Similarly, we adjust the perturbation radius $\rho$ based on prompt difficulty, thereby improving convergence stability. We provide details in Appendix~\ref{app:focal_weight}.

\paragraph{Gradient.}
We apply gradient centralization (GC)~\cite{gcsam} to both the ascent and descent steps of SAM, reducing gradient norm throughout optimization. Empirically, this improves the model's exploration capability during post-training. We provide details in Appendix~\ref{app:gradient_centralization}.

Algorithm \ref{alg:sagrpo} integrates these adaptations into \texttt{Flat}-RLVR built upon a GRPO baseline.

\section{Experiments}
\label{experiments}
\subsection{Experimental Setup}
\label{sec:exp_setup}
\paragraph{Training Setup.}
In our experiments, we adopt Qwen2.5-Math-1.5B~\cite{qwen25math} and Llama-3.2-1B-Instruct~\cite{llama} backbones and train them on the MATH dataset~\cite{hendrycksmath2021}, comprising 7.5K problems. Full training details are in Appendix~\ref{exper_details}.

\begin{table}[t]
  \caption{Solvable-problem coverage (\%) averaged over all in-domain benchmarks on Qwen2.5-Math-1.5B. Each row is a category defined by whether the base and the RL-tuned model solve a problem ($\checkmark$) or not ($\times$).}
  \centering
  \small
  \label{tab:cov_analysis}
  \begin{tabular}{cc|cc}
    \toprule
    \textbf{Base} & \textbf{RL-Tuned} & \textbf{GRPO} & \textbf{\texttt{Flat}-GRPO} \\
    \midrule
    $\checkmark$ & $\checkmark$ & $57.7$ & $ 61.2$ \\
    $\checkmark$ & $\times$     & $6.3$  & $2.8$  \\
    $\times$     & $\checkmark$ & $5.4$  & $6.3$  \\
    $\times$     & $\times$     & $30.6$ & $29.7$ \\
    \midrule
    \multicolumn{2}{c|}{\textbf{Total}} & $100.0$&$100.0$ \\
    \bottomrule
  \end{tabular}
  \vskip -0.1in
\end{table}

\begin{table*}
  \caption{
    Pass@$1$/Pass@256 ($\%$) values across two base models with varying RLVR methods at $G=6$ group size, reported as mean $\pm 95\%$ CI over 3 seeds.\textbf{Avg.}: average Pass@$1$/Pass@$256$. \textbf{Bold}: better within baseline/\texttt{Flat} pair.}
  \label{tab:generality_eval}
  \centering
  \resizebox{\textwidth}{!}{
  \begin{tabular}{l|l|c|ccccc}
    \toprule
    \textbf{Model}&\textbf{Method} & \textbf{Avg.} & \textbf{MATH500}    & \textbf{AMC23}           & \textbf{AIME24}              & \textbf{AIME25}             & \textbf{AIME26}\\
    \midrule
    \multirow{6}{*}{Qwen2.5} 
    & GRPO                &\textbf{15.6}\cib{1.0}/64.8\cib{3.7} &\textbf{34.6}\cib{0.0}/\textbf{72.7}\cib{0.7} &\textbf{30.2}\cib{2.8}/95.8\cib{7.1} & \textbf{7.0}\cib{1.2}/62.2\cib{4.7} & 2.9\cib{0.1}/48.9\cib{12.6} & 3.7\cib{0.8}/44.4\cib{4.7} \\
    & \texttt{Flat}-GRPO  &15.2\cio{0.6}/\textbf{67.3}\cio{3.0} &34.0\cio{0.0}/72.5\cio{0.5} &28.9\cio{2.4}/\textbf{96.7}\cio{7.1} &6.5\cio{1.5}/\textbf{64.4}\cio{9.5} &\textbf{3.3}\cio{2.4}/\textbf{51.1}\cio{4.7} &\textbf{3.8}\cio{0.7}/\textbf{52.2}\cio{4.7} \\
    \cmidrule{2-8}
    & DAPO                &16.3\cib{0.2}/65.2\cib{5.8} &35.7\cib{0.1}/72.1\cib{2.0} &32.0\cib{0.7}/\textbf{98.3}\cib{3.5} &6.9\cib{1.0}/63.3\cib{8.2} &3.1\cib{0.8}/51.1\cib{9.5}& 3.9\cib{0.2}/41.1\cib{12.6} \\
    & \texttt{Flat}-DAPO  &16.3\cio{0.1}/\textbf{68.6}\cio{0.5} &35.7\cio{0.2}/\textbf{72.4}\cio{0.4} &\textbf{32.1}\cio{0.2}/97.5\cio{6.2} &6.9\cio{0.6}\textbf{66.7}\cio{14.3} &3.1\cio{0.2}/\textbf{57.8}\cio{4.7} &3.9\cio{0.5}/\textbf{48.9}\cio{12.6} \\
    \cmidrule{2-8}
    & CISPO               &\textbf{15.7}\cib{0.3}/65.5\cib{2.6} & \textbf{34.7}\cib{0.1}/\textbf{72.8}\cib{1.4}& 30.6\cib{0.9}/98.3\cib{7.1}& \textbf{6.9}\cib{0.8}/63.3\cib{8.2} &2.9\cib{0.4}/50.0\cib{14.3} &3.9\cib{0.4}/43.3\cib{0.0} \\
    & \texttt{Flat}-CISPO &15.6\cio{0.1}/\textbf{68.5}\cio{3.0} &33.8\cio{0.2}/72.6\cio{0.4} &\textbf{30.7}\cio{0.8}/\textbf{99.2}\cio{3.5} &6.5\cio{0.8}/\textbf{67.8}\cio{4.7} &\textbf{3.1}\cio{0.4}/\textbf{55.6}\cio{17.2} &\textbf{4.2}\cio{0.2}/\textbf{47.8}\cio{4.7} \\
    \midrule       
    \multirow{6}{*}{Llama3.2} 
    & GRPO                &6.5\cib{0.2}/48.4\cib{0.4} &\textbf{18.8}\cib{0.6}/\textbf{66.1}\cib{2.0} &11.8\cib{0.5}/\textbf{90.8}\cib{7.1} &\textbf{1.5}\cib{0.3}/35.6\cib{12.6} &0.2\cib{0.1}/26.7\cib{8.2} &0.3\cib{0.1}/23.3\cib{0.0} \\ 
    & \texttt{Flat}-GRPO  &6.5\cio{0.1}/\textbf{50.4}\cio{2.9} &18.4\cio{0.3}/64.6\cio{1.7} &\textbf{12.3}\cio{0.1}/87.5\cio{6.2} &1.5\cio{0.2}/\textbf{41.1}\cio{9.5} &0.2\cio{0.0}/\textbf{33.3}\cio{14.3} &0.3\cio{0.1}/25.6\cio{4.7} \\
    \cmidrule{2-8}
    & DAPO                &\textbf{6.7}\cib{0.1}/47.9\cib{4.6} &\textbf{19.4}\cib{0.2}/\textbf{64.6}\cib{0.0} &\textbf{12.0}\cib{0.5}/88.3\cib{7.1} &\textbf{1.9}\cib{0.2}/38.9\cib{4.7} &0.1\cib{0.1}/23.3\cib{8.2} &\textbf{0.5}\cib{0.2}/\textbf{24.4}\cib{4.7} \\
    & \texttt{Flat}-DAPO  &6.4\cio{0.2}/\textbf{49.1}\cio{7.9} &18.5\cio{0.3}/64.5\cio{1.7} &11.8\cio{1.1}/\textbf{90.0}\cio{6.2} &1.3\cio{0.4}/\textbf{43.3}\cio{8.2} &\textbf{0.2}\cio{0.1}/\textbf{26.7}\cio{14.3} &0.4\cio{0.1}/21.1\cio{12.6} \\
    \cmidrule{2-8}
    & CISPO               &\textbf{6.7}\cib{0.5}/47.4\cib{3.4} &\textbf{20.8}\cib{0.0}/65.2\cib{1.4} &11.3\cib{2.8}/87.5\cib{6.2} &1.3\cib{0.3}/\textbf{40.0}\cib{14.3} &0.1\cib{0.1}/24.4\cib{4.7} &0.2\cib{0.1}/20.0\cib{8.2} \\
    & \texttt{Flat}-CISPO &6.6\cio{0.2}/\textbf{48.7}\cio{0.5} &19.5\cio{0.2}/\textbf{65.5}\cio{0.7} &\textbf{11.9}\cio{1.0}/\textbf{89.2}\cio{7.1} &\textbf{1.5}\cio{0.3}/38.9\cio{4.7} &\textbf{0.2}\cio{0.1}/\textbf{26.7}\cio{0.0} &\textbf{0.4}\cio{0.4}/\textbf{23.3}\cio{0.0} \\
    \bottomrule
  \end{tabular}
  }
  \vskip -0.1in
\end{table*}

\paragraph{Evaluation Setup.}
We evaluate on five in-domain mathematical reasoning benchmarks---MATH500~\cite{math500}, AMC23, AIME24~\cite{aime24}, AIME25~\cite{aime25}, and AIME26~\cite{aime26}---and report Pass@K curves traced from $K=1$ to $K=256$, where Pass@1 measures single-attempt accuracy and Pass@256 captures diversity-driven coverage of the solution space.  Detailed evaluation protocols are provided in Appendix~\ref{exper_details}.

\subsection{Reasoning Capacity Evaluation}
\label{sec:main_results}
Motivated by \citet{doesRL}, we adopt two analysis methods for evaluating \emph{Reasoning Capacity}: Pass@$K$ metrics and Solvable-Problem Coverage Analysis. Concretely, Pass@$K$ metrics measure the boundaries of reasoning ability by sampling $k$ outputs from the model. Solvable-Problem Coverage Analysis measures the fraction of problems solvable by the base model or the RL-tuned model, indicating whether \texttt{Flat}-RLVR truly mitigates the narrowing of reasoning capacity boundaries stemming from the naive RLVR framework.

\paragraph{Pass@K.} We evaluate the effect of applying SAM to GRPO, namely \texttt{Flat}-GRPO, with group size $G=6$. Figure~\ref{fig:main} shows Pass@$K$ curves on in-domain benchmarks, comparing the base model, GRPO, and \texttt{Flat}-GRPO from $K=1$ to $K=256$. At small $K$, both GRPO and \texttt{Flat}-GRPO outperform the base model across benchmarks, reflecting the well-known single-attempt gains of RLVR. As $K$ grows, however, GRPO's advantage erodes, and the base model eventually matches or even surpasses it---indicating that despite its higher Pass@$1$, GRPO yields a \emph{narrower} reasoning coverage than the base model, a signature of distribution sharpening. In contrast, \texttt{Flat}-GRPO maintains a consistent advantage over both GRPO and the base model across all values of $K$, indicating that SAM preserves the model's exploratory capacity while retaining the single-attempt gains of RLVR.

\paragraph{Solvable-Problem Coverage Analysis.} We analyze the set of solvable problems for the base model relative to GRPO and to \texttt{Flat}-GRPO. For each pair, we measure \textbf{i)} the fraction of problems solvable by the base model ($\checkmark$) but not by the RL-tuned model ($\times$)---i.e., problems the RL stage \emph{forgets}---and \textbf{ii)} the fraction newly solved by the RL-tuned model ($\checkmark$) but not by the base model ($\times$). This decomposition isolates how much reasoning coverage RLVR sacrifices in exchange for the new problems it acquires. Table~\ref{tab:cov_analysis} reports GRPO forgets $6.3\%$ of the problems originally solvable by the base model while newly solving only $5.4\%$---a net erosion of base-model coverage that is symptomatic of distribution sharpening. \texttt{Flat}-GRPO, however, substantially improves both sides: it forgets far \emph{fewer} problems ($2.8\%$) while solving \emph{more} new ones ($6.3\%$). SAM mitigates GRPO's tendency to collapse toward a narrow subset of the base model's solvable problems, thereby alleviating distribution sharpening.
Additional details are in Appendix~\ref{app:solvable_coverage}.

\begin{table}[t]
  \caption{Pass@1/Pass@256 ($\%$) for GRPO, F-GRPO, and \texttt{Flat}-GRPO (ours), with Qwen2.5-Math-1.5B (top) and Llama-3.2-1B-Instruct (bottom) as the backbone, reported as mean $\pm 95\%$ CI over 3 seeds.}
  \centering
  \resizebox{\columnwidth}{!}{
  \begin{tabular}{l|ccc}
    \toprule
    \textbf{Benchmarks} & \textbf{GRPO}               &\textbf{F-GRPO}              &\textbf{\texttt{Flat}-GRPO}            \\
    \midrule
    MATH500             & \textbf{34.6}\ci{0.0} / \textbf{72.7}\ci{0.3} & 33.2\cib{0.0} / 72.1\cib{0.6}                   & 34.0\cio{0.0} / 72.5\cio{0.2}                  \\
    AMC23               & \textbf{30.2}\ci{1.1} / 95.8\ci{2.9}          & 28.4\cib{1.8} / \textbf{98.3}\cib{1.4}          & 28.9\cio{1.0} / 96.7\cio{2.9}                  \\
    AIME24              & \textbf{7.0}\ci{0.5} / 62.2\ci{1.9}                    & 6.4\cib{0.7} / \textbf{66.7}\cib{6.7}                   & 6.5\cio{0.6} / 64.4\cio{3.8} \\
    AIME25              & 2.9\ci{0.1} / 48.9\ci{5.1}           & 2.6\cib{0.2} / 47.8\cib{1.9}                    & \textbf{3.3}\cio{1.0} / \textbf{51.1}\cio{1.9} \\
    AIME26              & 3.7\ci{0.3} / 44.4\ci{1.9}                    & 3.5\cib{0.1} / 41.1\cib{1.9}           & \textbf{3.8}\cio{0.3} / \textbf{52.2}\cio{1.9} \\
    \midrule
    Avg.                & \textbf{15.6}\ci{0.45} / 64.8\ci{1.7}          & 14.8\cib{0.7} / 65.2\cib{2.3}                   & 15.2\cio{0.4} / \textbf{67.3}\cio{1.3}         \\
    \midrule
    MATH500            & 18.8\ci{0.3} / 66.1\ci{0.8}                    & \textbf{20.7}\cib{0.1} / \textbf{66.4}\cib{0.2} & 18.4\cio{0.1} / 64.6\cio{0.7}                  \\
    AMC23              & 11.8\ci{0.2} / 90.8\ci{2.9}                    & 11.2\cib{0.2} / \textbf{94.3}\cib{1.6}          & \textbf{12.3}\cio{0.0} / 87.5\cio{2.5}         \\
    AIME24             & 1.5\ci{0.1} / 35.6\ci{5.1}                     & \textbf{2.4}\cib{0.1} / 37.8\cib{1.9}           & 1.5\cio{0.1} / \textbf{41.1}\cio{3.9}          \\
    AIME25             & 0.2\ci{0.0} / 26.7\ci{3.3}                     & 0.2\cib{0.0} / 23.3\cib{6.7}                    & 0.2\cio{0.0} / \textbf{33.3}\cio{5.8}          \\
    AIME26             & 0.3\ci{0.1} / 23.3\ci{0.0}                     & \textbf{0.5}\cib{0.0} / \textbf{30.0}\cib{5.8}           & 0.3\cio{0.0} / 25.6\cio{1.9}                   \\
    \midrule
    Avg.               & 6.5\ci{0.1} / 48.4\ci{2.0}                     & \textbf{6.9}\cib{0.1} / 50.3\cib{2.8}           & 6.5\cio{0.1} / \textbf{50.4}\cio{1.9}          \\
    \bottomrule
  \end{tabular}
  }
  \label{tab:sa_grpo_f_grpo}
  \vskip -0.1in
\end{table}

\begin{table*}[t]
    \caption{Pass@1/Pass@256 ($\%$) results on a larger model (3B) with mean ±95$\%$ CI over 3 seeds. \texttt{Flat}-RLVR consistently improves the average Pass@$256$ while preserving the average Pass@$1$ at this larger scale.}
    \label{tab:larger_scale}
    \centering
    \resizebox{\textwidth}{!}{
    \begin{tabular}{l|c|ccccc}
         \toprule
         \textbf{Method} & \textbf{Avg.} & \textbf{MATH500} & \textbf{AMC23} & \textbf{AIME24} & \textbf{AIME25} & \textbf{AIME26} \\
         \midrule
         BASE & 16.7\ci{0.2}/60.8\ci{1.3} & 44.5\ci{0.1}/70.5\ci{0.3} & 31.1\ci{0.8}/98.3\ci{7.2} & 4.5\ci{0.2}/47.8\ci{9.6} & 1.6\ci{0.1}/50.0\ci{8.3} & 2.0\ci{0.3}/37.8\ci{9.6} \\

         \midrule
         GRPO & \textbf{18.5}\cib{0.2}/59.4\cib{0.7} & \textbf{46.9}\cib{0.1}/\textbf{70.3}\cib{0.6} & \textbf{36.9}\cib{0.9}/95.8\cib{3.6} & \textbf{5.2}\cib{0.1}/42.2\cib{12.6} & 1.6\cib{0.5}/46.7\cib{8.3} & \textbf{2.3}\cib{0.3}/42.2\cib{4.8} \\

         \texttt{Flat}-GRPO & 18.2\cio{0.2}/\textbf{61.4}\cio{1.6} & 46.7\cio{0.1}/69.5\cio{0.3} & 35.9\cio{0.7}/\textbf{97.5}\cio{0.0} & 4.7\cio{0.5}/\textbf{48.9}\cio{9.6} & 1.6\cio{0.2}/\textbf{47.8}\cio{4.8} & 2.2\cio{0.3}/\textbf{43.3}\cio{0.0} \\
         \bottomrule
    \end{tabular}
    }
    \vskip -0.1in
\end{table*}

\begin{table}[t]
    \caption{Pass@1/Pass@256 ($\%$) with Qwen2.5-Coder-1.5B-Instruct on code generation test (MBPP), reported as mean ±95$\%$ CI over 3 seeds. \texttt{Flat}-GRPO achieves higher Pass@256 than GRPO at comparable Pass@1.}
    \centering
    \small
    \begin{tabular}{l|c}
         \toprule
         \textbf{Method} &\textbf{MBPP} \\
         \midrule
         BASE & 42.3\ci{0.1} / 91.5\ci{1.1} \\ 
         \midrule
         GRPO & 56.7\cib{0.0} / 85.6\cib{0.9} \\
         \texttt{Flat}-GRPO & 54.5\cio{0.1} / 89.6\cio{1.6}\\
         \bottomrule
    \end{tabular}
    \label{tab:code_gen}
    \vskip -0.1in
\end{table}

\subsection{Extended Evaluations}
We further evaluate \texttt{Flat}-RLVR from four perspectives: \textbf{(1) Generality evaluation} across various RLVR variants, including GRPO, DAPO, and CISPO. This evaluation examines whether flatness-seeking optimization transfers across distinct group-relative algorithms. We keep the group size $G{=}6$ and report Pass@$1$ / Pass@$256$; \textbf{(2) Comparative evaluation} against F-GRPO~\cite{fgrpo}, which addresses the same goal via difficulty-aware advantage scaling; \textbf{(3) Scalability Evaluation} with Larger Backbone on Qwen2.5-3B-Instruct~\cite{qwen2.5,qwen2025}, which examines whether the benefits of \texttt{Flat}-RLVR persist at a larger model scale; and \textbf{(4) Cross-Domain Evaluation} with Code Generation Test on Qwen2.5-Coder-1.5B-Instruct~\cite{qwen2_coder}, which examines whether the reasoning capacity benefit of \texttt{Flat}-RLVR extends beyond the mathematical domain. All experiments follow the identical training and evaluation protocol described in Appendix~\ref{exper_details}.

\paragraph{Generality Evaluation across GRPO variants.}
Table~\ref{tab:generality_eval} reports the following results: On Qwen2.5-Math-1.5B, flatness-seeking optimization improves the average Pass@256 by \textcolor{blue}{+2.5} (GRPO $\rightarrow$ \texttt{Flat}-GRPO), \textcolor{blue}{+3.4} (DAPO $\rightarrow$ \texttt{Flat}-DAPO), and \textcolor{blue}{+3.0} (CISPO $\rightarrow$ \texttt{Flat}-CISPO), while Pass@1 remains stable across all three; on Llama-3.2-1B-Instruct, the same comparisons yield \textcolor{blue}{+2.0}, \textcolor{blue}{+1.2}, and \textcolor{blue}{+1.3} on Pass@256, with Pass@1 preserved. Taken together, these results indicate that the diversity benefit of \texttt{Flat}-RLVR broadly transfers across distinct group-relative algorithms and backbone families.

\paragraph{Comparative Evaluation with Baseline.}
Table~\ref{tab:sa_grpo_f_grpo} reports that \texttt{Flat}-GRPO surpasses F-GRPO in average Pass@256 on both backbones---by \textcolor{blue}{+2.1} on Qwen2.5-Math-1.5B and \textcolor{blue}{+0.1} on Llama-3.2-1B-Instruct---while remaining competitive on Pass@1. This suggests that mitigating sharpening at the optimization-geometry level is at least as effective as F-GRPO's advantage-reweighting strategy, and complementary to it: these two methods operate on distinct levers, i.e., loss geometry by ours vs. advantage magnitude by F-GRPO, and our approach shows a competitive gain in the diversity-critical regime.

\paragraph{Scalability Evaluation with Larger Backbone.} To test whether the benefit of \texttt{Flat}-RLVR persists at a larger backbone scale, we train Qwen2.5-3B-Instruct with the same hyperparameters as in our main experiments. Table~\ref{tab:larger_scale} shows that \texttt{Flat}-GRPO improves Pass@256 by \textcolor{blue}{+2.0} over GRPO at a negligible Pass@1 cost (\textcolor{red}{-0.3}), suggesting that its benefits are not limited to small-scale models. 

\paragraph{Cross-Domain Evaluation with Code Generation Test.} We evaluate the model's sampling efficiency and reasoning capacity on Mostly Basic Programming Problems (MBPP) train/test split~\cite{mbpp} via Pass@K metric. Table~\ref{tab:code_gen} shows that \texttt{Flat}-GRPO achieves a comparable sampling efficiency gain to GRPO (\textcolor{red}{-2.2}) while substantially better preserving reasoning capacity (\textcolor{blue}{+4.0}). This indicates that the benefit of \texttt{Flat}-RVLR extends beyond mathematical reasoning to code generation, suggesting that flatness-aware optimization is broadly applicable across diverse domains. 

\begin{figure}[t]
  \vskip 0.2in
\centering
  \begin{subfigure}[t]{0.495\linewidth}
  \includegraphics[width=\linewidth]{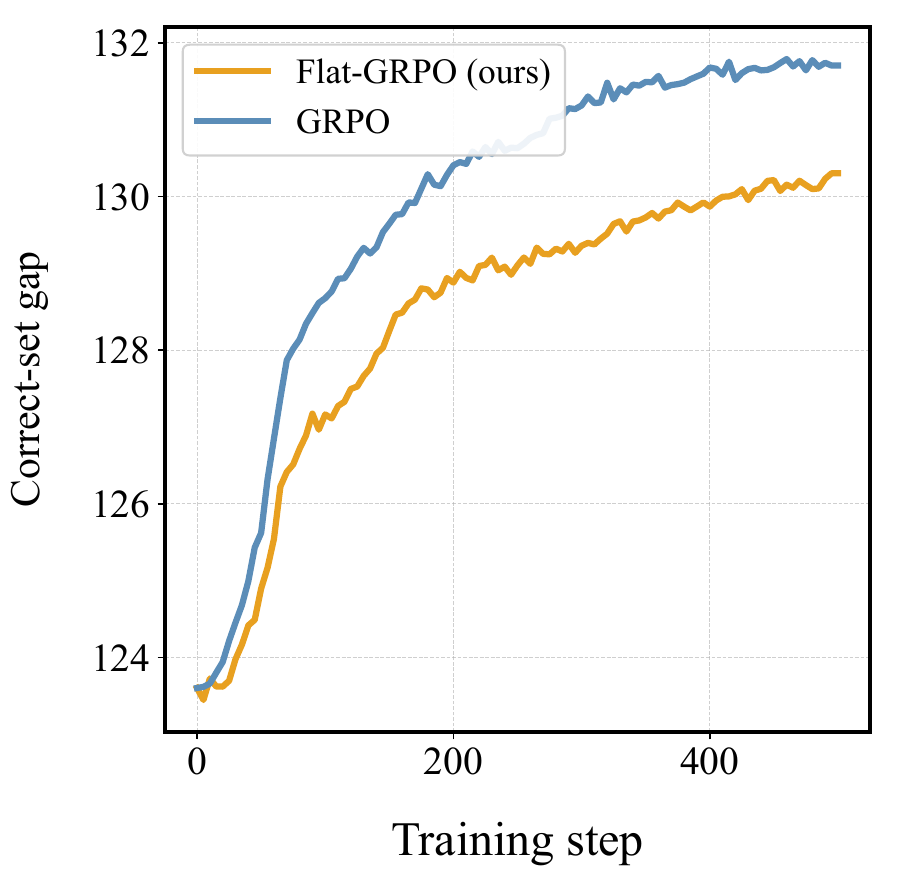}
    \caption{Correct-set gap $D_{h,l}$}
  \label{fig:correct_set_log}
\end{subfigure}
  \hfill
  \begin{subfigure}[t]{0.495\linewidth}
   \includegraphics[width=\linewidth]{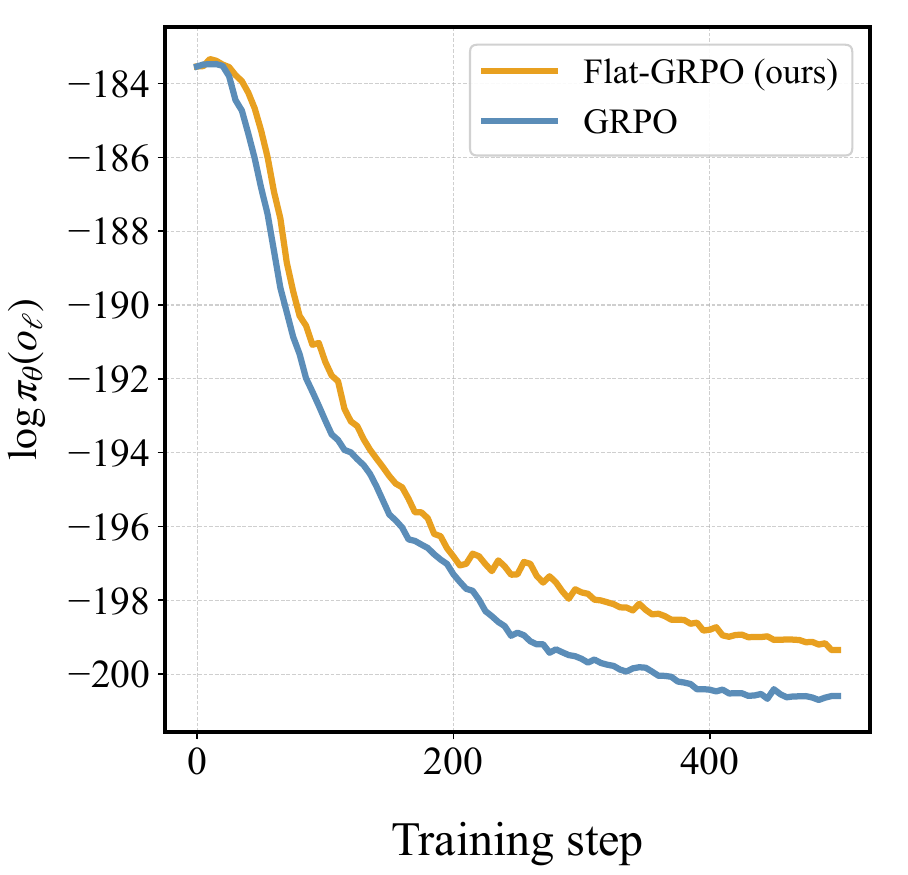}
    \caption{Trajectory of $\log \pi_\theta(o_\ell)$}
    \label{fig:correct_gap}
  \end{subfigure}
  \caption{Correct-set log-probability gap $D_{h,l}$ on a fixed probe set~(a) and trajectory of $\log \pi_\theta(o_\ell)$, the log-probability of the lowest-probability correct response~(b) during training.}
\end{figure}

\subsection{Empirical Validation of Theoretical Analysis}
\label{sec:empirical_validation}
To validate the theoretical analysis of \texttt{Flat}-RLVR's benefits, we further evaluate \texttt{Flat}-RLVR from three perspectives: \textbf{(1) Correct-set log probability gap}, which is a key theoretical quantity in Definition~\ref{def2}; \textbf{(2) Rare-correct log-probability}, which demonstrates that \texttt{Flat}-RLVR mitigates distribution sharpening by preserving rare-correct trajectories; and \textbf{iii) Low-pass filter (LPF)}~\cite{LPF} of Qwen2.5-Math-1.5B, where lower values indicate a flatter minimum. Full training details are in Appendix~\ref{exper_details}.

\paragraph{Correct-set log probability gap.}
We trace the correct-set log probability gap $D$ training Qwen2.5-Math-1.5B on MATH dataset. The correct-set log probability gap $D=\log \pi_\theta (o_{h}|q) - \log \pi_\theta(o_l|q)$ (referred to Definition~\ref{def2}) indicates the difference between a high-probability correct-output $o_h$ and low-probability correct-output $o_l$. Reducing this gap growth is the core mechanism by which \texttt{Flat}-RLVR mitigates distribution sharpening.

Specifically, \textbf{i)} for a given prompt, we measure $o_h$ (high-probability correct output) and $o_l$ (low-probability correct output) before training. \textbf{ii)} After one-step training, we measure them to compute the changes of the gap $D$ and keep track of it during training. Figure~\ref{fig:correct_set_log} shows that \texttt{Flat}-GRPO reduces the gap of correct-set log probability compared with GRPO during training, indicating \texttt{Flat}-RLVR mitigates the correct-set distribution sharpening.

\paragraph{Rare-correct log-probability.}
We simultaneously trace the mean probability mass assigned to the fixed rare-correct trajectories $o_l$ throughout the same training process. Figure~\ref{fig:correct_gap} shows that \texttt{Flat}-GRPO retains more probability mass on rare correct trajectories $(o_l)$ than GRPO, which suggests that it helps preserve the model's reasoning capacity by keeping rare correct solution alive.

\begin{table}[t]
    \centering
    \caption{Low-Pass Filter (LPF) of the model fine-tunded by GRPO and \texttt{Flat}-GRPO algorithms at perturbation scale $\sigma = 0.001$.}
    \begin{tabular}{l|c}
         \toprule
         \textbf{Method} & \textbf{LPF $\downarrow$} \\
         \midrule
         GRPO & 0.4434 \\
         \textbf{\texttt{Flat}-GRPO} & \textbf{0.0598} \\
         \bottomrule
    \end{tabular}
    \label{tab:LPF}
  \vskip -0.1in
\end{table}

\paragraph{Low-pass filter (LPF).}
In addition, to verify that \texttt{Flat}-GRPO really find the flatter region than GRPO, we measure the Low-Pass Filter (LPF) flatness defined as the expected increase in loss under Gaussian parameter perburbaions: 
\begin{equation}
    \text{LPF}(\theta) = \mathbb{E}_{\delta\sim\mathcal{N}(0,\sigma^2)}[\mathcal{L}(\theta+\delta)] - \mathcal{L}(\theta).
\end{equation}

We construct a fixed evaluation batch of 32 prompts from the model’s training dataset (MATH dataset). For each prompt, we sample $G=6$ rollouts from the model and compute advantages $\hat{A}_i = (r_i - \bar{r}) / \sigma_r$, defining $\mathcal{L}$ as the GRPO loss on this fixed batch. The identical batch and perturbation seeds are used for both models to controlled comparison. Perturbation are filter-normalized scaling the noise on each parameter by its norm. 

Table~\ref{tab:LPF} shows the model trained with \texttt{Flat}-GRPO converges to a flatter region than GRPO. This is consistent with our theoretical claim that flatness-aware optimization mitigates distribution sharpening.

\subsection{Analysis of Cost-Benefit Trade-off}
\label{exp:periodic_sam}
We further investigate how applying SAM to the GRPO baseline affects performance and incurs additional time and memory overhead, as a function of the SAM period $n$, which controls how often the SAM ascent-descent step is invoked. We treat the period $n$ as the key factor that governs the practical value of \texttt{Flat}-RLVR. All experiments use Qwen2.5-Math-1.5B as the backbone and follow the configuration. We measure all cost values with the same computing resource for a fair comparison.

\begin{figure}[ht]
  \vskip 0.2in
\centering
  \begin{subfigure}[t]{0.495\linewidth}
  \includegraphics[width=\linewidth]{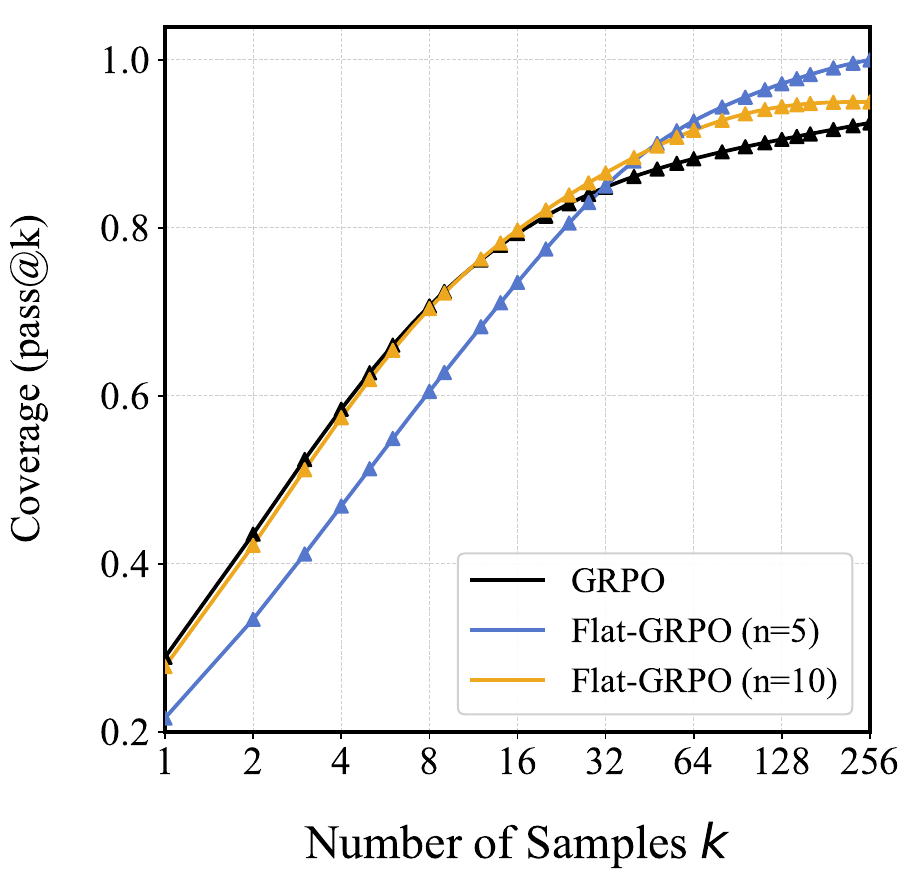}
  \caption{AMC23}
\end{subfigure}
  \hfill
  \begin{subfigure}[t]{0.495\linewidth}
   \includegraphics[width=\linewidth]{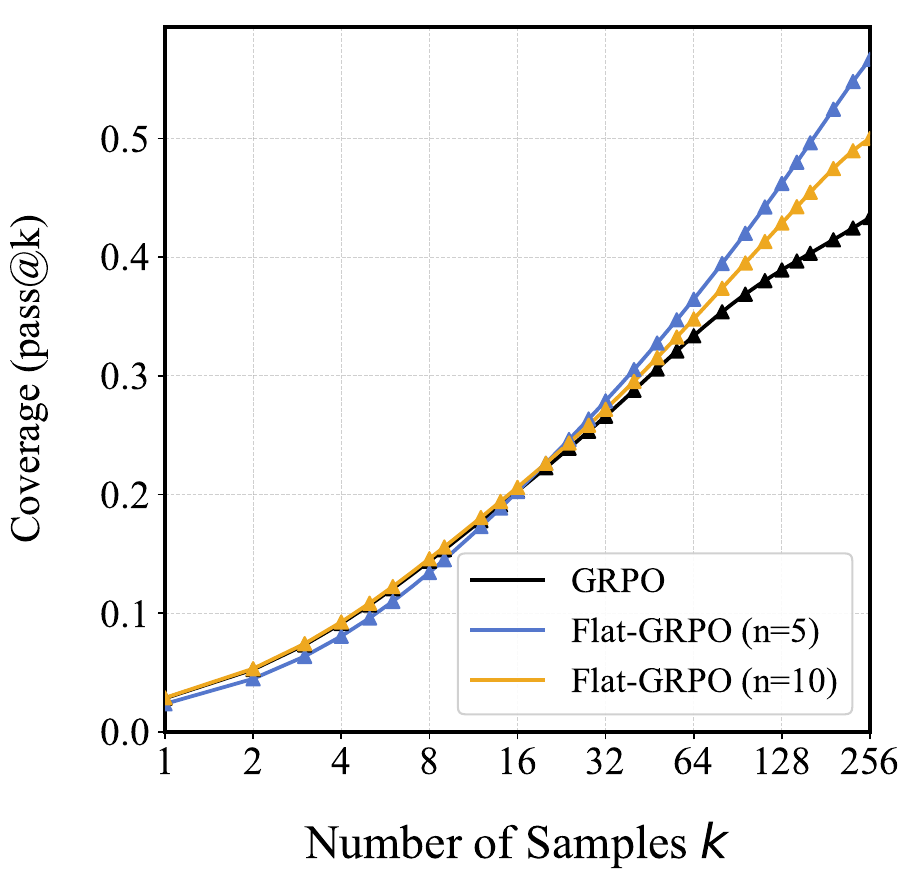}
    \caption{AIME25}
    \label{fig:sam_l_2}
  \end{subfigure}
  \caption{\label{fig:sam_k}Pass@K curves on (a) AMC23 and (b) AIME25 test datasets (from $K=1$ to $K=256$) comparing GRPO with \texttt{Flat}-GRPO under two SAM periods, $n=5$ and $n=10$.}
\end{figure}

\paragraph{Performance.}
\label{app:sam_k_performance}
Figure~\ref{fig:sam_k} compares GRPO against periodic SAM with $n \in \{5, 10\}$ on AMC23 and AIME25 test datasets, plotting Pass@K from $K=1$ to $K=256$. The figure reveals a clear tradeoff: shortening the SAM application interval (from $n=10$ to $n=5$) substantially boosts Pass@K as $K$ grows, but at the cost of a non-trivial drop in single-attempt accuracy (Pass@1). 

\paragraph{Computational Overhead.}
\label{app:sam_k_time}
Table~\ref{tab:compute_scale} reports the total wall-clock time and the average per-GPU memory allocation in the training process. Across both periods, the total training time stays within $0.1$ hours of the GRPO baseline and the per-GPU memory remains comparable or lower. The results indicate that periodic application of SAM incurs negligible additional training time and memory usage relative to the baseline, guaranteeing the scalability of the \texttt{Flat}-RLVR.

\begin{table}[t]
  \centering
  \small
  \caption{Wall-clock training time and averaged per-GPU memory allocation for \texttt{Flat}-GRPO with $n$-step}
  \label{tab:compute_scale}
  \begin{tabular}{l|c|c}
    \toprule
    \textbf{Method} & \textbf{Total (h)} & \textbf{allocated($\%$)}\\
    \midrule
    \texttt{Flat}-GRPO ($n=5$)  & $17.00$ & $90.82$\\
    \texttt{Flat}-GRPO ($n=10$) & $16.87$ & $88.42$ \\
    GRPO                        & $16.91$ & $90.96$ \\
    \bottomrule
  \end{tabular}
\vskip -0.1in
\end{table}

\begin{table}[t]
    \centering
    \small
    \caption{Average time for each step in \texttt{Flat}-GRPO. We compare SAM steps with ordinary steps to estimate the computational overhead stemming from the additional gradient of SAM.}
    \begin{tabular}{l|c}
         \toprule
         &  \textbf{per-step wall clock time (s)} \\
         \midrule
         ordinary step & 90.17 \\
         \textbf{SAM step} & \textbf{90.56 (+0.43$\%$)} \\
         \bottomrule
    \end{tabular}
    \label{tab:sam_step_time}
    \vskip -0.1in
\end{table}

Furthermore, to quantify the cost of additional gradient computation from SAM, we examine the per-step wall-clock time of Flat-GRPO on Qwen2.5-3B-Instruct. Table~\ref{tab:sam_step_time} shows that a SAM step takes only 0.39s $(+0.43\%)$ more than an ordinary step, confirming that the additional gradient computation remains negligible at scale.

\section{Conclusion}
In this work, we present a way to mitigate the distribution sharpening in group-relative RLVR, by utilizing flatness-seeking optimization. Concretely, we provide a mathematical analysis of the one-step parameter update in RLVR, showing that flat-minima search dampens the abrupt increase in probability, thereby mitigating distribution sharpness of LLM post-training. We propose the \texttt{Flat}-RLVR framework by integrating sharpness-aware minimization into RLVR. 
Throughout experiments, \texttt{Flat}-RLVR implemented as \texttt{Flat}-GRPO/DAPO/CISPO mitigates the distribution sharpening induced by RLVR, consistently improving the diversity of the output distribution while preserving accuracy with our LLM post-training adaptations. These imply flatness serves as the central catalyst to mitigate the distribution sharpening.

\section*{Acknowledgments}
This work was supported by the National Research Foundation of Korea (NRF) grant funded by the Korea government (MSIT) (No. RS-2024-00459023), (No. RS-2026-25541562, AI-Based Dynamic Risk Assessment Platform for Accelerating SMR and Advanced Reactor Design), and the Institute of Information $\&$ Communications Technology Planning $\&$ Evaluation (IITP) grant funded by the Korean Government (MSIT) (No. RS-2020-II201336, Artificial Intelligence Graduate School Program at UNIST), (No. RS2025-25442824, AI Star Fellowship Program (UNIST)), (No. RS-2026-25528781, Hyper-scale Industrial AI Research Support (R$\&$D) Program, Development of an industry-specified intelligent data processing and federated learning platform), (No. IITP-2026-RS-2022-00156361, Innovative Human Resource Development for Local Intellectualization program). This work was also supported by the grant of the Korea Health Technology R$\&$D Project through the Korea Health Industry Development Institute (KHIDI), funded by the Ministry of Health $\&$ Welfare, Republic of Korea (No. RS-2025-02223382).

\bibliography{main}
\bibliographystyle{icml2026}

\newpage
\clearpage

\appendix

\section{Analysis on Flatness}

\subsection{Proof of Lemma~\ref{lem:Flat-RLVR}}
\label{app:proof_lemma1}
For ease of derivation, we begin with a simplified, sequence-level form of the GRPO objective per query in Eq~\eqref{eq:grpo_objective}:
\begin{equation}
    \mathcal{L}(\theta) = -\frac{1}{G}\sum^{G}_{i=1}\gamma_{i}(\theta)\,\hat{\mathcal{A}}_{i}, 
    \gamma_{i}(\theta) = \frac{\pi_{\theta}(o_{i}\mid q)}{\pi_{\theta_{\mathrm{old}}}(o_{i}\mid q)}.
\label{eq:grpo_loss}
\end{equation}
Since $\nabla_{\theta}\gamma_{i}(\theta) = \gamma_{i}(\theta)s_{i,t}$ where $ s_{i,t} = \nabla_{\theta}\log\pi_{\theta}(o_{i}\mid q)$, the gradient of $\mathcal{L}$ with respect to $\theta$ is
\begin{equation}
    \nabla_{\theta}\mathcal{L}(\theta) \\= -\frac{1}{G}\sum^{G}_{i=1}\hat{\mathcal{A}}_{i}\,\gamma_{i}(\theta)s_{i, t}.
\end{equation}
Applying the product rule, the Hessian of $\gamma_i$ is
\begin{equation}
    \nabla^{2}_{\theta}\gamma_{i}(\theta) = \gamma_{i}(\theta)(s_{i,t}s_{i,t}^{\top} + \nabla^{2}_{\theta}\log\pi_{\theta}(o_i | q)).
\end{equation}

\paragraph{The gradient and Hessian at $\theta_t$.}
Under the one-step update assumption $\pi_{\theta} \approx \pi_{\theta_{\mathrm{old}}}$ (equivalently, $\gamma_{i}(\theta) \approx 1$), the gradient and Hessian at $\theta_t$ simplify to
\begin{equation}
    \nabla_{\theta}\mathcal{L}(\theta_t) \approx -\frac{1}{G}\sum^{G}_{i=1}\hat{\mathcal{A}}_{i}s_{i,t} = -g_t,
\end{equation}
\begin{multline}
    \nabla^{2}_{\theta}\mathcal{L}(\theta_t) \approx -\frac{1}{G}\sum^{G}_{i=1} \hat{\mathcal{A}}_{i}\!(s_{i,t}s_{i,t}^{\top} + \nabla^{2}_{\theta}\log\pi_{\theta}(o_i|q))\\ = -(M_t + B_t).
\end{multline}
where we identify $g_t, M_t$, and $B_t$ as defined in Lemma~\ref{lem:Flat-RLVR}.

\paragraph{The worst-case perturbation.}
Following Eq.~\eqref{eq:sam_perturbation}, the worst-case perturbation is
\begin{equation}
    \epsilon^{*} = \rho\cdot\frac{\nabla_{\theta}\mathcal{L}(\theta_t)}{\lVert\nabla_{\theta}\mathcal{L}(\theta_t)\rVert_2} = -\rho\cdot\frac{g_t}{\lVert g_t \rVert_2}.
\end{equation}

\paragraph{The first-order Taylor expansion.}
The gradient of the perturbed loss is approximated by a first-order Taylor expansion~\cite{sam}:
\begin{equation}
    \nabla_{\theta}\mathcal{L}(\theta_t+\epsilon^{*}) \approx \nabla_{\theta}\mathcal{L}(\theta_t) + \nabla^{2}_{\theta}\mathcal{L}(\theta_t)\,\epsilon^{*} + R_T.
\end{equation}
where $R_T$ collects higher-order Taylor residuals.

\paragraph{The one-step update for \texttt{Flat}-RLVR.}
The \texttt{Flat}-RLVR update minimizes $-\mathcal{L}(\theta_t + \epsilon^*)$, giving
\begin{multline}
    \theta^{\text{Flat}}_{t+1} \\
    = \theta_t - \alpha\Big[(-g_t) + (-(M_t + B_t))\Big(-\rho\frac{g_t}{\lVert g_t \rVert_2}\Big)\Big] - \alpha R_T \\
    = \theta_t + \alpha\Big[g_t - (M_t + B_t)\rho\frac{g_t}{\lVert g_t\rVert_2}\Big] - \alpha R_T \\
    = \theta_t + \alpha \Big(g_t - \rho \frac{M_tg_t}{\lVert g_t \rVert_2}\Big) - \alpha \rho \frac{B_tg_t}{\lVert g_t \rVert _2} - \alpha R_T,
\end{multline}
which recovers Eq~\eqref{eq:lemma_decomp}.

\subsection{Proof of Lemma~\ref{lem:correct_gap}}
\label{app:proof_lemma2}
By definition, the correct-support log-probability gap is
\begin{equation}
    D_{h,l}(\theta; q) 
    = \log\pi_\theta(o_h \mid q) - \log\pi_\theta(o_l \mid q).
\end{equation}

\paragraph{First-order approximation of the gap change.}
For a one-step parameter update $\Delta\theta_t = \theta_{t+1} - \theta_t$, a first-order Taylor expansion around $\theta_t$ gives
\begin{multline}
    D_{h,l}(\theta_t + \Delta\theta_t; q) 
    \\ \approx D_{h,l}(\theta_t; q) 
      + \nabla_\theta D_{h,l}(\theta_t; q)^\top \Delta\theta_t.
\end{multline}
Since $\nabla_\theta \log\pi_\theta(o_i \mid q) | _{\theta_t} = s_{i,t}$, we have
\begin{equation}
    \nabla_\theta D_{h,l}(\theta_t; q) 
    = s_{h,t} - s_{l,t} =: \Delta s_t,
\end{equation}
and therefore
\begin{multline}
    \mathfrak{D}_t(\Delta\theta_t) 
    \\:= D_{h,l}(\theta_t + \Delta\theta_t; q) - D_{h,l}(\theta_t; q)
    \approx \Delta s_t^\top \Delta\theta_t,
\end{multline}
with recovers Eq~\eqref{eq:first_order_correct_set}.

\paragraph{Gap change under the GRPO update.}
The standard GRPO update is $\Delta\theta_t^{\mathrm{GRPO}} = \alpha g_t$, so
\begin{equation}
    \label{eq:grpo_update_term}
    \mathfrak{D}_t(\Delta \theta_t^{\mathrm{GRPO}}) \approx \Delta s_t^\top (\alpha g_t) = \alpha \Delta s_t^\top g_t.
\end{equation}

\paragraph{Gap change under the \texttt{Flat}-RLVR update.}
For Lemma~\ref{lem:Flat-RLVR}, the \texttt{Flat}-RLVR update is
\begin{equation}
    \Delta\theta_t^{\mathrm{Flat}} 
    = \alpha\!\left(g_t - \rho\,\frac{M_t g_t}{\|g_t\|_2}\right) + R_t,
\end{equation}
where $R_t = -\alpha\rho\, B_t g_t / \|g_t\|_2 - \alpha R_T$ is the residual.
Substituting into the first-order gap approximation:
\begin{multline}
        \mathfrak{D}_t(\Delta\theta_t^{\mathrm{Flat}})
    \approx \Delta s_t^\top \Delta\theta_t^{\mathrm{Flat}}  \\
    = \Delta s_t^\top \!\left[\alpha g_t 
       - \alpha\rho\,\frac{M_t g_t}{\|g_t\|_2} + R_t\right]  \\
    = \alpha\,\Delta s_t^\top g_t 
       - \alpha\rho\,\frac{\Delta s_t^\top M_t g_t}{\|g_t\|_2} 
       + \Delta s_t^\top R_t.
\end{multline}

\paragraph{Identifying the decomposition.}
Recognizing the GRPO term from Eq~\eqref{eq:grpo_update_term}, and defining
\begin{equation}
    C_t := \frac{\Delta s_t^\top M_t g_t}{\|g_t\|_2}, \, r_t := \Delta s_t^\top R_t,
\end{equation}
we obtain
\begin{equation}
    \mathfrak D_t(\Delta\theta_t^{\texttt{Flat}}) = \mathfrak D_t(\Delta\theta_t^{\mathrm{GRPO}}) - \alpha\rho C_t + r_t,
\end{equation}
which recovers Eq~\eqref{eq:lemma_final_eq}.

\subsection{Proof of Proposition~\ref{prop:mitigation}}
\label{app:proof_proposition}
\paragraph{Pointwise gap difference.}
By the correct-set Gap Decomposition Lemma~\ref{lem:correct_gap}, for each step $t$ the difference between the \texttt{Flat}-RLVR-induced and GRPO-induced correct-set gap changes satisfies
\begin{equation}
    \mathfrak{D}_t(\Delta\theta_t^{\mathrm{Flat}}) - \mathfrak{D}_t(\Delta\theta_t^{\mathrm{GRPO}}) = -\alpha\rho C_t + r_t,
\end{equation}
where
\begin{equation}
    C_t = \frac{\Delta s_t^\top M_t g_t}{\|g_t\|_2}, \qquad r_t = \Delta s_t^\top R_t.
\end{equation}

\paragraph{Summation over the trajectory.}
Summing the pointwise relation over $t = 0, 1, \ldots, T-1$ yields
\begin{multline}
    \sum_{t=0}^{T-1}\mathfrak{D}_t(\Delta\theta_t^{\mathrm{Flat}}) \\
    = \sum_{t=0}^{T-1}\mathfrak{D}_t(\Delta\theta_t^{\mathrm{GRPO}}) 
    - \sum_{t=0}^{T-1}\alpha\rho C_t 
    + \sum_{t=0}^{T-1} r_t.
\end{multline}

\paragraph{Applying the dominance condition.}
By the assumption of the proposition,
\begin{equation}
    \sum_{t=0}^{T-1}\alpha\rho C_t \;\ge\; \sum_{t=0}^{T-1} r_t,
\end{equation}
so the net correction $-\sum_{t}\alpha\rho C_t + \sum_{t} r_t \le 0$. Substituting back gives
\begin{equation}
    \sum_{t=0}^{T-1}\mathfrak{D}_t(\Delta\theta_t^{\mathrm{Flat}}) 
    \;\le\; 
    \sum_{t=0}^{T-1}\mathfrak{D}_t(\Delta\theta_t^{\mathrm{GRPO}}),
\end{equation}
which is the claimed inequality.

\section{Adaptations for LLM post-training}
\subsection{Focal Weight}
\label{app:focal_weight}
Inspired by F-GRPO~\cite{fgrpo}, which decays the advantage via a focal weight conditioned on prompt difficulty, we adapt this idea to SAM's hyperparameter $\rho$ to promote stable training. Specifically, given $X$ correct rollouts out of a total of $N$, we define the focal weight as
\begin{equation}
    p_{\mathrm{mean}} = \frac{X}{N}.
\end{equation}
We then modulate SAM's perturbation radius $\rho$ accordingly:
\begin{equation}
    \rho = \rho \cdot\bigl(p_{\mathrm{mean}}\bigr)^{\gamma}.
\end{equation}

This formulation suppresses the influence of perturbations when the success rate is low (\emph{high difficulty})---thereby preventing unstable updates---while preserving the perturbation signal needed to discover rare correct trajectories when the success rate is high.

In practice, the focal weight itself is computed from the per-batch success rate,
\begin{equation}
    \hat{p}_{\mathrm{mean}} = \frac{\hat{X}}{\hat{N}},
\end{equation}
where $\hat{N}$ is the training batch size and $\hat{X}$ is the number of correct samples within the batch.

\subsection{Gradient Centralization}
\begin{figure}[t]
  \vskip 0.2in
  \includegraphics[width=\columnwidth]{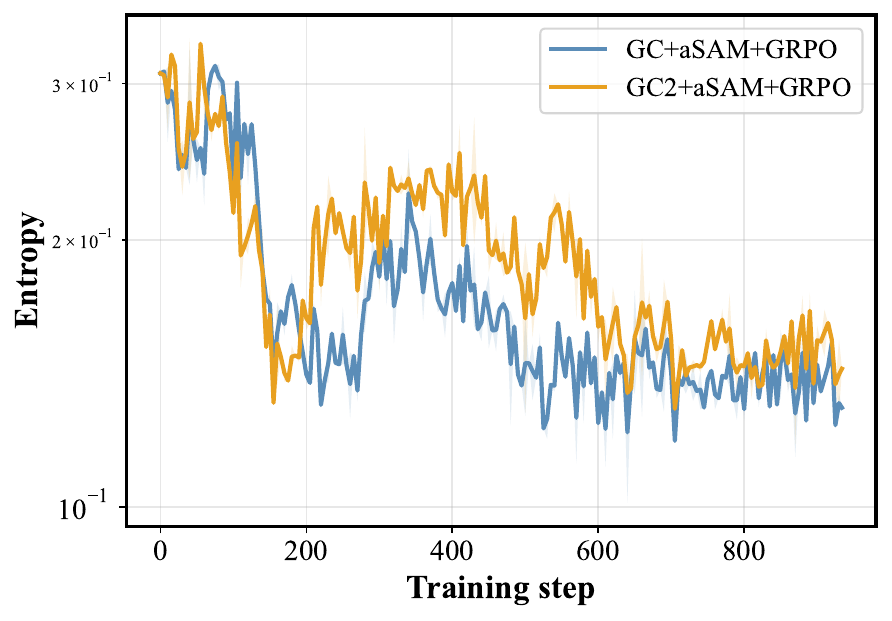}
  \caption{Training entropy curves on GSM8K with Qwen2.5-0.5B-Instruct: SAM with Gradient Centralization (GC) applied at both the ascent and descent steps (GC2+aSAM+GRPO) versus only at the ascent step (GC+aSAM+GRPO). Applying GC before the descent step substantially mitigates entropy collapse compared to the ascent-only variant.}
  \label{fig:gc}
\end{figure}

\label{app:gradient_centralization}
Applying Gradient Centralization before SAM's \emph{ascent step} is known to reduce gradient variance, thereby suppressing noise and improving the quality of sharpness estimation~\cite{gcsam}. This property is particularly well suited to the group-relative RLVR setting, where sparse rewards and high-variance advantage estimates are inherent~\cite{skip_grpo,tree_grpo}. We therefore apply Gradient Centralization before the ascent step, providing a more reliable perturbation direction for SAM in the RLVR setting.

Specifically, let the weight matrix of a particular layer be $W \in \mathbb{R}^{M \times N}$, and let $w_i$ denote the $i$-th row vector of $W$. Then we can define the following:
\begin{equation}
    \nabla_{GC} J(w_i) = \nabla J(w_i) - \mu,
\end{equation}
where $\mu = \frac{1}{n}\sum_{j=0}^{n-1}\nabla J(w_{ij})$ and $J$ is the objective function.

In addition, we further apply Gradient Centralization before SAM's \emph{descent step} to reduce the gradient norm. \citet{GC} shows that the norm of the projected gradient $\nabla_{\mathrm{GC}}$ admits the following bound:
\begin{multline}
    \lVert\nabla_{\mathrm{GC}} J(W)\rVert^{2}_{2} = \bigl(P\,\nabla J(W)\bigr)^{\top} P\,\nabla J(W) \\
    = \nabla J(W)^{\top}\bigl(I - ee^{\top}\bigr)\nabla J(W) \\
    = \lVert\nabla J(W)\rVert^{2}_{2} - \lVert e^{\top}\nabla J(W)\rVert^{2}_{2} \;\leq\; \lVert\nabla J(W)\rVert^{2}_{2}.
\end{multline}
where $P = I - ee^{\top}$ is the projection matrix and $e = \tfrac{1}{\sqrt{n}}\mathbf{1} \in \mathbb{R}^{n}$ is the unit vector with all components equal.

Reducing the gradient norm is known to bias the optimizer toward flat minima, thereby improving model generalization~\cite{reduce_gradient_norm}. When combined with SAM, this property further enhances the model's exploration ability: as shown in Figure~\ref{fig:gc}, applying GC before the descent step yields substantially less entropy collapse than the variant without it. Note that both variants already applied SAM-$n$ and Cosine Annealing. Full training details are deferred to Appendix~\ref{exper_details}.

\subsection{Annealing SAM $\rho$}
\label{app:annealing}
\begin{figure}[t]
  \vskip 0.2in
  \includegraphics[width=\columnwidth]{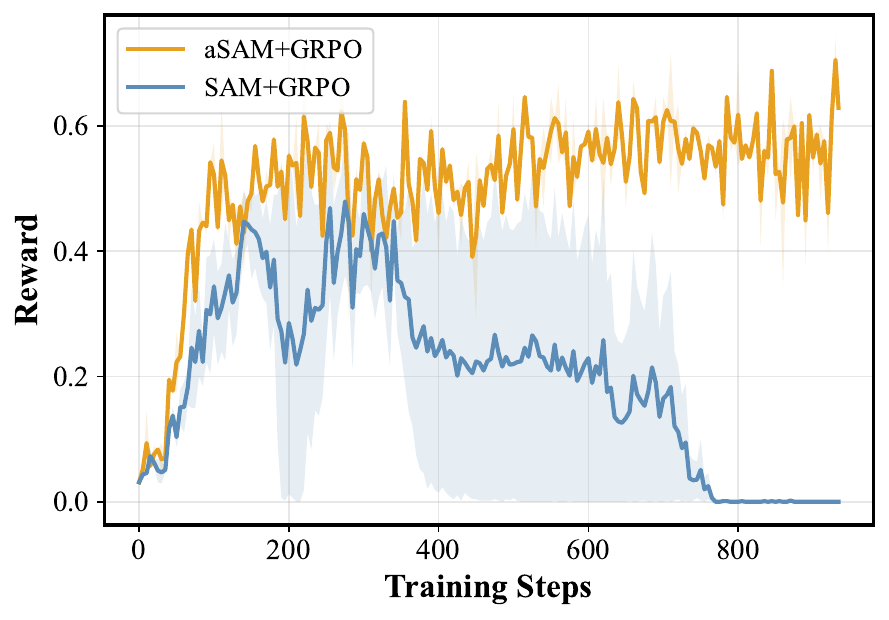}
  \caption{Training reward curves on GSM8K with Qwen2.5-0.5B-Instruct: SAM with cosine annealing (aSAM + GRPO) versus without (SAM + GRPO) (i.e., periodic-SAM without cosine annealing). Without annealing, the training reward collapses to zero, halting effective learning.}
  \label{fig:annealing}
\end{figure}

For training stability, we adopt a cosine annealing schedule~\cite{cosine} that gradually decays SAM's hyperparameter $\rho$ over the course of training. Concretely, $\rho$ at step $t$ follows
\begin{equation}
    \rho_{t} = \rho_{\mathrm{min}} + \tfrac{1}{2}\,\bigl(\rho_{\mathrm{init}} - \rho_{\mathrm{min}}\bigr)\!\left(1 + \cos\!\left(\frac{T_{\mathrm{cur}}}{T}\pi\right)\right),
\end{equation}
where $\rho_{\mathrm{init}}$ and $\rho_{\mathrm{min}}$ denote the initial and minimum values of $\rho$, respectively, and $T_{\mathrm{cur}}$ and $T$ denote the current and total numbers of training steps. By smoothly decaying $\rho$ as training progresses, cosine annealing gradually attenuates the influence of the perturbation. In the early phase of training, this prevents entropy collapse driven by overfitting to the training data, while in the later phase it promotes stable convergence.

Figure~\ref{fig:annealing} illustrates why this annealing is essential: without it, the training reward collapses to zero midway through training, halting effective learning. Gradually decaying $\rho$ is thus necessary for stable convergence in the later training phase. Full training details are deferred to Appendix~\ref{exper_details}.

\section{Ablation Study}
\label{app:ablation_study}
This appendix quantifies the individual contribution of each strategy introduced in Section~\ref{methods}. We conduct a systematic ablation on Qwen2.5-Math-1.5B with same environments of main results. Only difference stems from the application of each strategy. Full training details are in Appendix~\ref{exper_details}.

\begin{table*}[t]
    \caption{Additive Ablation Study on mathemathic benchmarks. We fix periodic calculation (SAM) and add each strategy separately; Cosine Annealing (Anne.), Gradient Centralization (GC), and Focal Weight (F), reporting Pass@1/Pass@256 at each stage. \textbf{Avg.:} average Pass@1/Pass@256. \du{} $\&$ \dd{}: the change Pass@1/Pass@256 relative to the naive GRPO.}
    \label{tab:additive_ablation}
    \centering
    \small
    \begin{tabular}{l|c|ccccc}
         \toprule
         \textbf{Method} & \textbf{Avg.} & \textbf{MATH500} & \textbf{AMC23} & \textbf{AIME24} & \textbf{AIME25} & \textbf{AIME26} \\
         \midrule
         GRPO & 15.2/63.1 & 34.6/73.0 & 28.8/92.5 & 6.4/63.3 & 2.8/43.3 & 3.3/43.3 \\
         \midrule
         GRPO + SAM & 9.8\dd{5.4}/64.6\du{1.5} & 22.5/73.0 & 18.8/100 & 4.2/63.3 & 1.6/46.7 & 2.1/40.0 \\
         GRPO + SAM + Anne. & 13.8\dd{1.4}/65.5\du{2.4} & 30.7/72.4 & 26.7/95.0 & 5.8/60.0 & 2.4/53.3 & 3.3/46.7 \\
         GRPO + SAM + GC & 11.6\dd{3.6}/64.7\du{1.6} & 25.8/72.8 & 22.5/97.5 & 4.4/56.7 & 2.3/50.0 & 3.1/46.7 \\ 
         GRPO + SAM + F & 15.5\du{0.3}/65.8\du{2.7} & 35.1/71.6 & 28.8/97.5 & 7.2/66.7 & 2.6/46.7 & 3.9/46.7 \\
         \midrule
         \texttt{Flat}-GRPO & 15.0\dd{0.2}/67.5\du{4.4} & 33.9/72.4 & 27.7/95.0 & 7.1/66.7 & 2.8/50.0 & 3.4/53.3 \\
         \bottomrule
    \end{tabular}
  \vskip -0.1in
\end{table*}

Table~\ref{tab:additive_ablation} reveals a clear gain per each strategy:
\begin{itemize}
    \item Just naively applicating SAM in RLVR can improve reasoning capacity (\textcolor{blue}{+1.5}), but sacrificing much of accuracy (\textcolor{red}{-5.4}) compared with GRPO.
    \item Cosine Annealing strategy gradually decays the SAM’s perturbation radius, which make training process more stable in terms of accuracy (\textcolor{blue}{+4.0}) compared with GRPO+SAM.
    \item Gradient Centralization strategy subtracts the gradient norm before ascent/descent step, thereby promoting stable (\textcolor{blue}{+1.8}) and diversity (\textcolor{blue}{+0.1}) compared with GRPO+SAM.
    \item Focal Weight strategy decays the radius of perturbation according to the success rate in group. It promotes stable training process (\textcolor{blue}{+5.7}) reducing the amount of radius if success rate is low and diversity (\textcolor{blue}{+1.2}) almost not decaying if success rate is high compared with GRPO+SAM.
\end{itemize}

In short, flatness-aware optimization can work itself for improving pass@k reasoning coverage, requiring some strategies for stablizing training process and amplifying its effect to some extent.

\section{The lightweight approximation for Memory Efficiency.}
\label{app:appro}
We adopt a lightweight approximation for applying SAM in RLVR; cache only the scalar norm of the original gradient, recompute a second perturbation $\epsilon^*$ at the perturbed parameters $\theta + \epsilon$. Concretely, by the definition of SAM, the two perturbations computed during this procedure are
\begin{equation}
    \epsilon = \rho \, \frac{\nabla J(\theta)}{\lVert\nabla J(\theta)\rVert}, \qquad
    \epsilon^{*} = \rho \, \frac{\nabla J(\theta+\epsilon)}{\lVert\nabla J(\theta)\rVert}.
\end{equation}
A first-order Taylor expansion of the gradient at the perturbed point yields
\begin{equation}
    \nabla J(\theta + \epsilon) = \nabla J(\theta) + H\,\epsilon + \mathcal{O}(\rho^{2}) \approx \nabla J(\theta) + \mathcal{O}(\rho),
\end{equation}
where $H := \nabla^{2} J(\theta)$ is the Hessian. Substituting this expansion into the expression for $\epsilon^{*}$ gives
\begin{equation}
    \epsilon^{*} \approx \epsilon + \mathcal{O}(\rho^{2}) \approx \epsilon.
\end{equation}

Hence, for sufficiently small $\rho$ (e.g., $\rho \leq 0.01$), this approximation is tight, and SAM can be applied by caching only a single scalar---virtually no additional spatial cost.

\section{Further Results}
\subsection{All Pass@K curves}
\begin{figure*}[ht]
  \vskip 0.2in
  \centering
  \includegraphics[width=\linewidth]{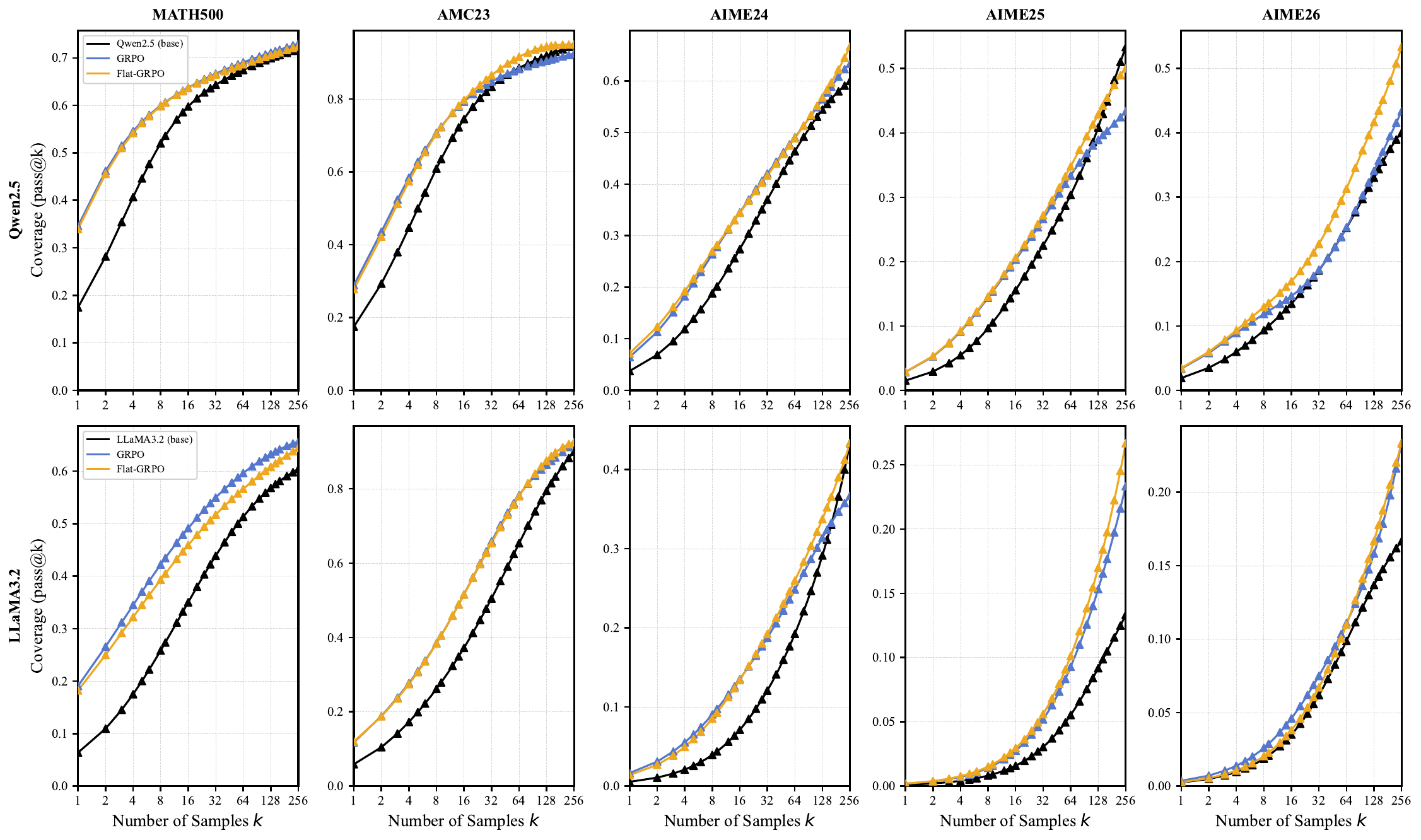}
  \caption{Pass@K curves on all in-domain benchmarks (MATH500, AMC23, AIME24, AIME25, AIME26), comparing the base model, GRPO, and \texttt{Flat}-GRPO (ours) from $K=1$ to $K=256$. Top row: Qwen2.5-Math-1.5B; bottom row: Llama-3.2-1B-Instruct.}
  \label{fig:all_curves}
\end{figure*}

Figure~\ref{fig:all_curves} reports the per-benchmark Pass@K curves for all five benchmarks (MATH500, AMC23, AIME24, AIME25, AIME26) on both backbones, from $K=1$ to $K=256$, complementing the aggregated results in Section~\ref{sec:main_results}.

\subsection{Solvable-problem coverage}
\label{app:solvable_coverage}
For each problem, we utilize the per-problem correctness rate computed during the Pass@K evaluation in Section~\ref{sec:main_results}, where each problem is sampled $n=256$ times. We define a problem as \emph{solvable} by a given model if at least one of its $n$ sampled completions is correct. For each base--RL pair, every problem is then assigned to one of four categories according to whether it is solvable by the base model and by the RL-tuned model: solvable by both, by the base model only (\emph{forgotten}), by the RL-tuned model only (\emph{newly solved}), or by neither. The reported percentages are the fraction of problems in each category, averaged across all benchmarks.

\begin{table*}[h]
  \centering
  \small
  \caption{Full per-benchmark solvable-problem coverage (\%) for the base--RL pairs on both backbones. Each row is a category defined by whether the base model and the RL-tuned model solve a problem ($\checkmark$) or not ($\times$): ($\checkmark$/$\times$) counts problems the RL stage \emph{forgets}, and ($\times$/$\checkmark$) counts problems it \emph{newly solves}. Averages correspond to Table~\ref{tab:cov_analysis} in the main text.}
  \label{app:total_results}
  \begin{tabular}{l|cc|cccccc}
    \toprule
    \textbf{Backbone} & \textbf{Base} & \textbf{RL} & \textbf{MATH500} & \textbf{AMC23} & \textbf{AIME24} & \textbf{AIME25} & \textbf{AIME26} & \textbf{Avg.} \\
    \midrule
    \multicolumn{9}{l}{\textit{Qwen2.5-Math-1.5B \quad (Base vs.\ GRPO)}} \\
    \midrule
    Qwen2.5 & $\checkmark$ & $\checkmark$ & $71.2$ & $87.5$ & $53.3$ & $43.3$ & $33.3$ & $57.7$ \\
            & $\checkmark$ & $\times$     & $0.6$  & $7.5$  & $6.7$  & $10.0$ & $6.7$  & $6.3$  \\
            & $\times$     & $\checkmark$ & $1.8$  & $5.0$  & $10.0$ & $0.0$  & $10.0$ & $5.4$  \\
            & $\times$     & $\times$     & $26.4$ & $0.0$  & $30.0$ & $46.7$ & $50.0$ & $30.6$ \\
    \midrule
    \multicolumn{9}{l}{\textit{Qwen2.5-Math-1.5B \quad (Base vs.\ \texttt{Flat}-GRPO)}} \\
    \midrule
    Qwen2.5 & $\checkmark$ & $\checkmark$ & $70.8$ & $95.0$ & $56.7$ & $43.3$ & $40.0$ & $61.2$ \\
            & $\checkmark$ & $\times$     & $1.0$  & $0.0$  & $3.3$  & $10.0$ & $0.0$  & $2.8$  \\
            & $\times$     & $\checkmark$ & $1.6$  & $0.0$  & $10.0$ & $6.7$  & $13.3$ & $6.3$  \\
            & $\times$     & $\times$     & $26.6$ & $5.0$  & $30.0$ & $40.0$ & $46.7$ & $29.7$ \\
    \midrule
    \multicolumn{9}{l}{\textit{LLaMA-3.2-1B-Instruct \quad (Base vs.\ GRPO)}} \\
    \midrule
    LLaMA3.2 & $\checkmark$ & $\checkmark$ & $58.4$ & $87.5$ & $26.7$ & $3.3$  & $13.3$ & $37.8$ \\
             & $\checkmark$ & $\times$     & $2.0$  & $2.5$  & $16.7$ & $10.0$ & $3.3$  & $6.9$  \\
             & $\times$     & $\checkmark$ & $7.2$  & $5.0$  & $10.0$ & $20.0$ & $10.0$ & $10.4$ \\
             & $\times$     & $\times$     & $32.4$ & $5.0$  & $46.7$ & $66.7$ & $73.3$ & $44.8$ \\
    \midrule
    \multicolumn{9}{l}{\textit{LLaMA-3.2-1B-Instruct \quad (Base vs.\ \texttt{Flat}-GRPO)}} \\
    \midrule
    LLaMA3.2 & $\checkmark$ & $\checkmark$ & $56.4$ & $87.5$ & $30.0$ & $6.7$  & $13.3$ & $38.8$ \\
             & $\checkmark$ & $\times$     & $4.0$  & $2.5$  & $13.3$ & $6.7$  & $3.3$  & $6.0$  \\
             & $\times$     & $\checkmark$ & $8.0$  & $5.0$  & $13.3$ & $20.0$ & $10.0$ & $11.3$ \\
             & $\times$     & $\times$     & $31.6$ & $5.0$  & $43.3$ & $66.7$ & $73.3$ & $44.0$ \\
    \bottomrule
  \end{tabular}
  \vskip -0.1in
\end{table*}

\section{Experimental details}
\label{exper_details}

\subsection{Full Training Details}
\label{training_details_15B}
All methods are trained with a token-level loss function, and the GRPO variants are implemented following the official guidelines of the TRL framework~\cite{trl}. Qwen2.5-Math-1.5B is trained on $5\times$ NVIDIA Quadro RTX 8000 GPUs; Llama-3.2-1B-Instruct on a single Quadro RTX 8000; Qwen2.5-0.5B-Instruct on a single NVIDIA GeForce RTX 5070 Ti; Qwen2.5-3B-Instruct and Qwen2.5-Coder-1.5B-Instruct on a single NVIDIA A100. The full set of hyperparameters is summarized in Table~\ref{tab:hyperparams}.

\begin{table}[t]
  \centering
  \small
  \caption{Training hyperparameters for \texttt{Flat}-RLVR.}
  \begin{tabular}{l|lc}
    \toprule
    \textbf{Method}        &\textbf{Parameter}     & \textbf{Value}   \\
    \midrule
    \textbf{SAM-$n$}     &     period $n$            & $10$        \\
    \midrule
    \textbf{Cosine Annealing} &    $\rho_{\text{init}}$               & $0.01$  \\
                     & $\rho_{\text{min}}$                & $0.002$            \\
    \midrule
    \textbf{Focal Weight}   &   $\gamma$                           & $1.0$      \\
                            &    $\rho^{\text{focal}}_{\text{min}}$             & $0.001$     \\
    \bottomrule
  \end{tabular}
  \label{tab:SA_GRPO}
  \vskip -0.1in
\end{table}

\subsection{Full Training Details: \texttt{Flat}-RLVR}
\label{training_details_sagrpo}
For training the models in \texttt{Flat}-RLVR Framework, we decay the SAM hyperparameter $\rho$ and use the lightweight approximation for memory efficiency. Additional details are in Appendix~\ref{app:appro} and Appendix~\ref{app:annealing}. The full set of hyperparameters for \texttt{Flat}-RLVR is summarized in Table~\ref{tab:SA_GRPO}.

Note that $\rho^{\text{focal}}_{\mathrm{\text{min}}}$ acts as a lower bound on the perturbation radius after focal-weight modulation, ensuring that the perturbation retains a non-trivial effect even when the focal weight becomes small. 

\subsection{Full Training Details: F-GRPO}
We implement F-GRPO following the official formulation of \citet{fgrpo}. For each prompt $x$, we compute the empirical success rate $\hat{\mu}_{\mathrm{pos}}(x)$ over its group of rollouts and scale the group-relative advantage by the Focal weight $g(x) = (1 - \hat{\mu}_{\mathrm{pos}}(x))^{\gamma}$, which down-weights updates on high-success prompts to preserve solution diversity. Following the original work, we set $\gamma = 1$ and keep all other hyperparameters identical to the GRPO baseline.

\subsection{Evaluation Protocol}
\label{pass_protocol}
We compute the Pass@K metric following~\citet{pass}:
\begin{equation}
    \mathrm{Pass@}k \;:=\; \mathbb{E}_{\,\mathrm{Problems}}\!\left[1 - \frac{\binom{n-c}{k}}{\binom{n}{k}}\right],
\end{equation}
where $n$ is the total number of generated samples per problem and $c$ is the number of correct samples among them. We set $n=256$ for all benchmarks and report from Pass@$1$ to Pass@$256$. For MATH500 and AMC23, we set the completion length to $1024$ tokens, matching the training-time setting; for the AIME family of benchmarks, we extend it to $2048$ tokens to accommodate the longer reasoning chains required by these problems. For GSM8K, we use a completion length of $192$ tokens, again matching training. 

Across all benchmarks, we use a single, uniform system prompt: \textit{``Please reason step by step, and put your final answer within \texttt{\textbackslash boxed\{\}}.''}

\subsection{Licenses}

\paragraph{Datasets.}
\textbf{GSM8k}~\cite{gsm8k} is released under the MIT License. \textbf{MATH}~\cite{hendrycksmath2021} and its 500-problem subset \textbf{MATH500}~\citep{math500} are released under the MIT License. 
\textbf{AIME24}, \textbf{AIME25}, and \textbf{AIME26} are sourced from \texttt{math-ai/aime24\footnote{\url{https://huggingface.co/datasets/math-ai/aime24}}}~\citep{aime24}, \texttt{math-ai/aime25\footnote{\url{https://huggingface.co/datasets/math-ai/aime25}}}~\citep{aime25}, and \texttt{math-ai/aime26\footnote{\url{https://huggingface.co/datasets/math-ai/aime26}}}~\citep{aime26}, released under the Apache License 2.0.
\textbf{AMC23} is sourced from \texttt{math-ai/amc23\footnote{\url{https://huggingface.co/datasets/math-ai/amc23}}}; no explicit license is specified for these datasets. \textbf{MBPP} is sourced from \texttt{google-research/mbpp}\footnote{\url{https://huggingface.co/datasets/claudios/google-research-datasets__mbpp}}~\citep{mbpp}; no explicit license is specified for these datasets.

\paragraph{Models.}
\textbf{Qwen2.5-Math-1.5B}, \textbf{Qwen2.5-0.5B-Instruct}, and \textbf{Qwen2.5-Coder-1.5B-Instruct}~\citep{qwen2_coder, qwen2025} are released under the Apache License 2.0. \textbf{Qwen2.5-3B-Instruct}~\citep{qwen2.5, qwen2025} is released under the Qwen Research License\footnote{\url{https://huggingface.co/Qwen/Qwen2.5-3B-Instruct/blob/main/LICENSE}}. 
\textbf{Llama-3.2-1B-Instruct}~\citep{llama} is released under the Llama 3.2 Community License\footnote{\url{https://www.llama.com/llama3_2/license/}}.

\begin{table*}[t]
  \caption{Full training hyperparameters for all models and methods.}
  \label{tab:hyperparams}
  \centering
  \small
  \resizebox{\linewidth}{!}{
  \begin{tabular}{l|ccccc}
    \toprule
    \textbf{Parameter} & \textbf{Qwen2.5-Math-1.5B} & \textbf{Llama-3.2-1B-Instruct} & \textbf{Qwen2.5-0.5B-Instruct} & \textbf{Qwen2.5-3B-Instruct} & \textbf{Qwen2.5-Coder-1.5B-Instruct} \\
    \midrule
    Optimizer              & \multicolumn{5}{c}{AdamW} \\
    $(\beta_1, \beta_2)$   & \multicolumn{5}{c}{$(0.9, 0.999)$} \\
    Weight decay           & \multicolumn{5}{c}{$0.01$} \\
    Learning rate          & \multicolumn{5}{c}{$1\times10^{-6}$} \\
    LR scheduler           & \multicolumn{5}{c}{Linear} \\
    Num.\ generations ($G$)& \multicolumn{5}{c}{$6$} \\
    PPO epochs             & \multicolumn{5}{c}{$1$} \\
    KL coefficient         & \multicolumn{5}{c}{$0$} \\
    Sampling temperature   & \multicolumn{5}{c}{$1.0$} \\
    (top-$p$, top-$k$)     & \multicolumn{5}{c}{$(1.0, 0)$} \\
    \midrule
    Global batch size      & $180$  & $72$   & $96$  & $180$  & $18$  \\
    Grad.\ accum.\ steps   & $9$    & $18$   & $6$   & $9$    & $18$  \\
    Per-device batch size  & $4$    & $4$    & $16$  & $4$    & $1$   \\
    Training epochs        & $2$    & $3$    & $2$   & $2$    & $5$   \\
    Max length             & $1024$ & $1024$ & $192$ & $1024$ & $512$ \\
    \bottomrule
  \end{tabular}
  }
  \vskip -0.1in
\end{table*}


\end{document}